\documentclass[letterpaper]{article}
\usepackage{etoolbox}
\usepackage{arxiv}
\usepackage{times}
\usepackage{helvet}
\usepackage{courier}
\pdfoutput=1
\usepackage{amsfonts}
\usepackage{amssymb}
\usepackage{subcaption}
\usepackage{standalone}
\usepackage{color}
\usepackage{color, colortbl}
\usepackage{multirow}
\definecolor{high}{rgb}{0.7,.85,.85}
\usepackage[table]{xcolor}
\usepackage{siunitx}
\definecolor{lightgray}{gray}{0.9}
\usepackage{microtype}
\usepackage{pifont}

\usepackage{framed}
\usepackage{mathtools,xparse}
\usepackage{tikz}
\usepackage{pgfplots}
\usepackage[hidelinks]{hyperref}
\usepackage{booktabs}
\usepackage{textcomp}
\usepackage[numbers,sort&compress]{natbib}
\usepackage{algorithmicx}
\usepackage{algorithm}
\usepackage{algcompatible}
\usepackage{paralist}
\usepackage{enumitem}

\usepackage[skip=4pt, font=small]{caption}

\DeclarePairedDelimiter{\norm}{\lVert}{\rVert}
\NewDocumentCommand{\normL}{ s O{} m }{%
  \IfBooleanTF{#1}{\norm*{#3}}{\norm[#2]{#3}}_{L_2(\Omega)}%
}

\definecolor{ourgreen}{rgb}{0.0,0.5,0.0}
\newcommand{\comment}[1]{}

\newcommand{\freq}{\emph{Frequency}}
\newcommand{\syn}{\emph{Syntactic}}
\newcommand{\ngrams}{Google Books Ngram Corpus}
\newcommand{\dist}{\textsc{geodist}}
\newcommand{\hwplotred}{\raisebox{2pt}{\tikz{\draw[red,solid,line width=2pt](0,0) -- (5mm,0);}}}
\newcommand{\hwplotgreen}{\raisebox{2pt}{\tikz{\draw[green,solid,line width=2pt](0,0) -- (5mm,0);}}}

\renewcommand{\footnotesize}{\scriptsize}

\begin{document}
\newtoggle{anonymize}
\togglefalse{anonymize}
\title{Freshman or Fresher? Quantifying the Geographic Variation of Language in Online Social Media}
\iftoggle{anonymize} {
\author{ICWSM-16 Paper}
}{
\author{Vivek Kulkarni 
\and
Bryan Perozzi
\and
Steven Skiena \\
Stony Brook University \\
Department of Computer Science, USA \\
\{vvkulkarni,bperozzi,skiena@cs.stonybrook.edu\}
}
}
\maketitle

\begin{abstract}
In this paper we present a new computational technique to detect and analyze statistically significant geographic variation in language. 
While previous approaches have primarily focused on lexical variation between regions, our method identifies words that demonstrate semantic and syntactic variation as well.
Our meta-analysis approach captures statistical properties of word usage across geographical regions and uses statistical methods to identify significant changes specific to regions.

We extend recently developed techniques for neural language models to learn word representations which capture differing semantics across geographical regions.
In order to quantify this variation and ensure robust detection of true regional differences, we formulate a null model to determine whether observed changes are statistically significant.
Our method is the first such approach to explicitly account for random variation due to chance while detecting regional variation in word meaning.

To validate our model, we study and analyze two different massive online data sets: millions of tweets from Twitter spanning not only four different countries but also fifty states, as well as millions of phrases contained in the Google Book Ngrams.
Our analysis reveals interesting facets of language change at multiple scales of geographic resolution -- from neighboring states to distant continents.

Finally, using our model, we propose a measure of semantic distance between languages.
Our analysis of British and American English over a period of $100$ years reveals that semantic variation between these dialects is shrinking.
\end{abstract}

\vspace{-0.1in}

\section{Introduction}
\label{sec:Introduction}

\begin{figure}[ht!]
\centering
\includegraphics[trim = 1in 0.7in 1in 0.7in, clip, width=0.9\columnwidth]{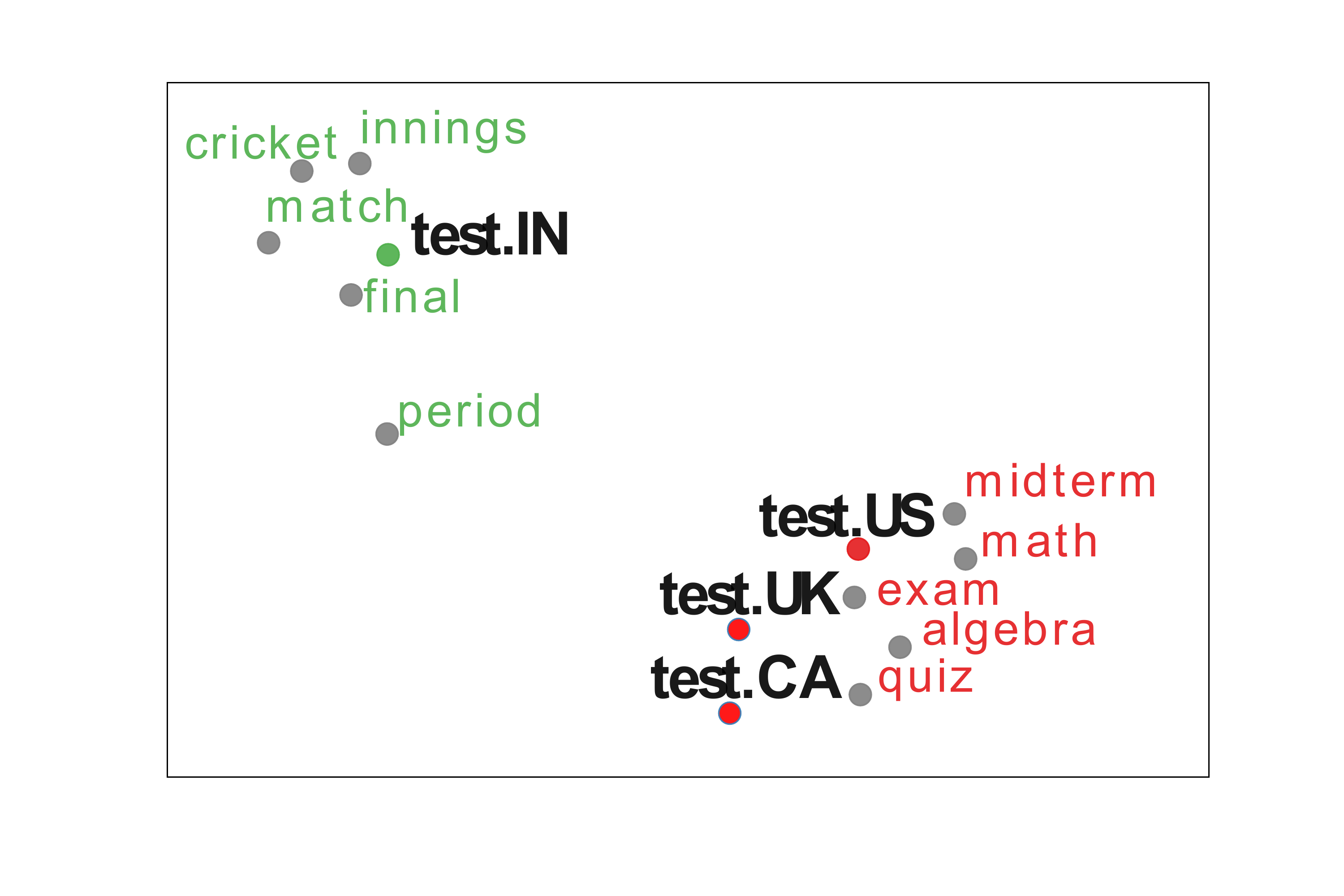}
\vspace{-0.05in}
\caption{The latent semantic space captured by our method (\dist) reveals geographic variation between language speakers.  
In the majority of the English speaking world (e.g.\ US, UK, and Canada) a \texttt{test} is primarily used to refer to an \texttt{exam}, while in India a \texttt{test}  indicates a lengthy cricket match which is played over five consecutive days.}
\label{fig:cw}
\end{figure}

\begin{figure*}[t!]
	\vspace{-0.05in}
	\begin{center}
	\begin{subfigure}{\columnwidth}
		\includegraphics[width=0.95\columnwidth]{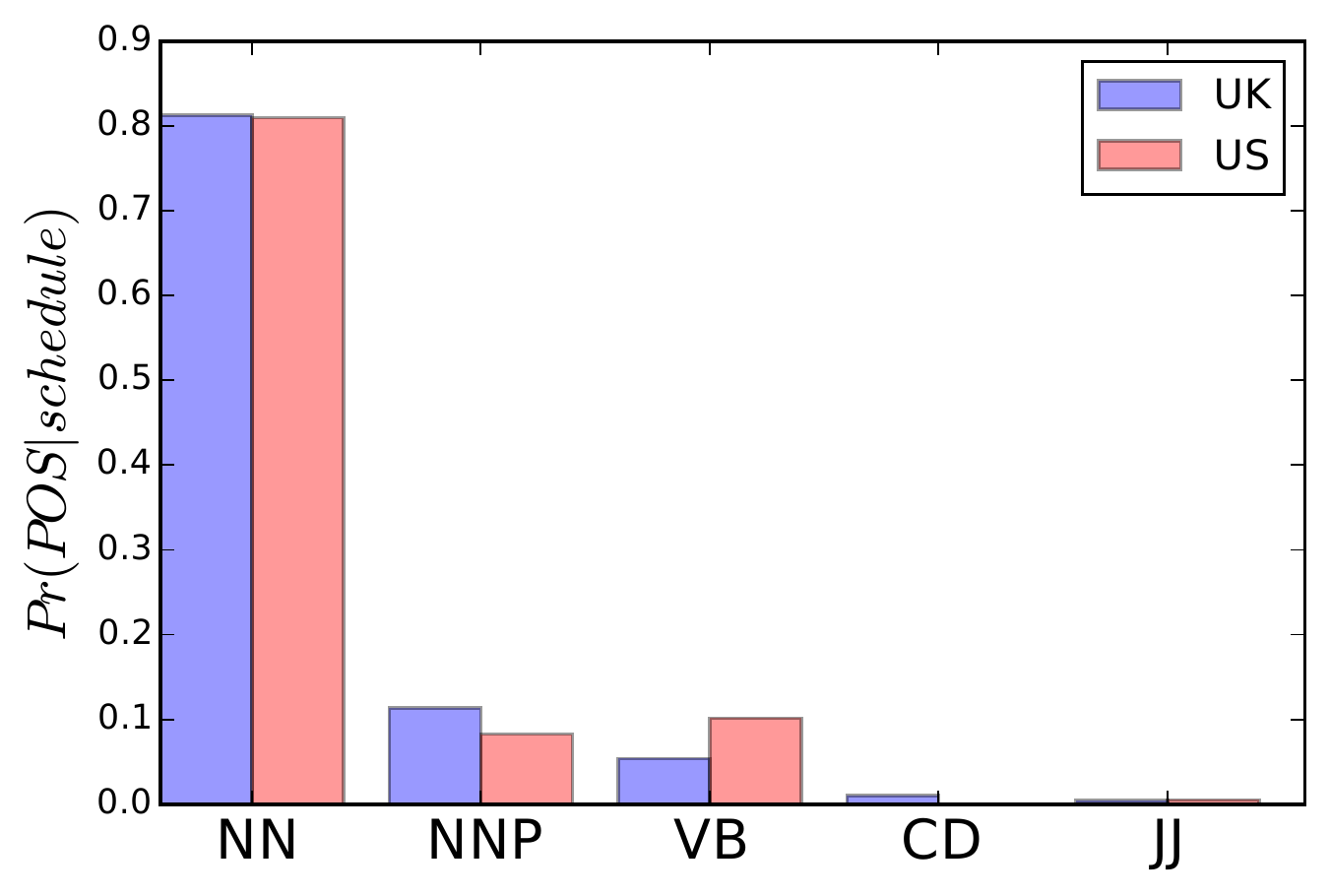}
		\caption{Part of Speech distribution for \texttt{schedule} (\syn)}
		\label{fig:schedule_pos}
	\end{subfigure}
	\begin{subfigure}{\columnwidth}
		\includegraphics[trim = 1in 0.3in 1.0in 0.5in, clip, width=0.9\columnwidth]{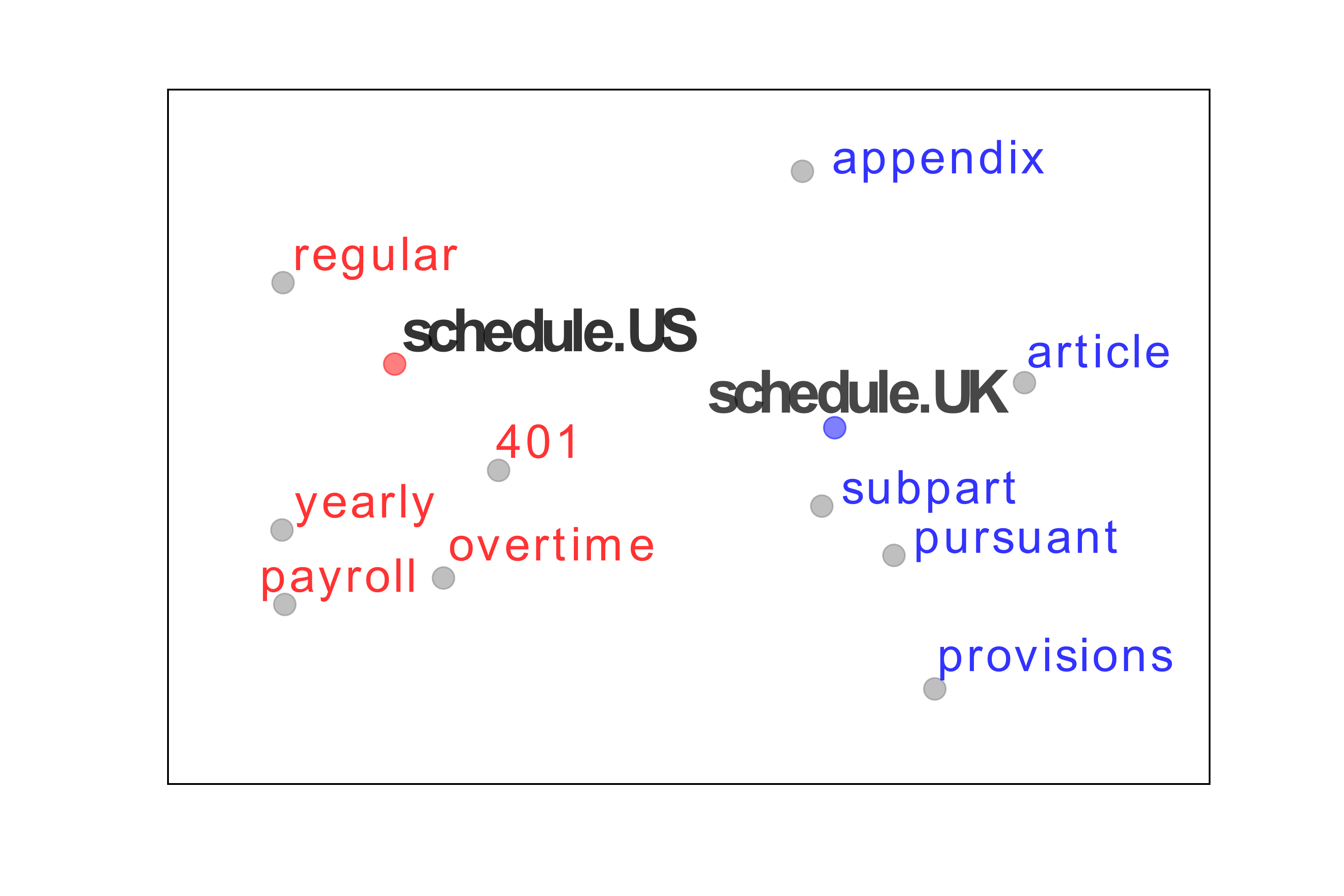}
		\caption{Latent semantic space captured by \dist\ method.}
		\label{fig:schedule_vis}
	\end{subfigure}
	\end{center}
	\vspace{-0.05in}
	\caption{The word \texttt{schedule} differs in its semantic usage between US and UK English which \dist\ (see Figure \ref{fig:schedule_vis}) detects. While \texttt{schedule} in the USA refers to a \emph{``scheduling time''}, in the UK \texttt{schedule} also has the meaning of an \emph{``addendum to a text''}. However the \syn\ method (see Figure \ref{fig:schedule_pos}) does not detect this semantic change since \texttt{schedule} is dominantly used as a noun (NN) in both UK and the USA.}
	\label{fig:crownjewel}	
\end{figure*}

Detecting and analyzing regional variation in language is central to the field of socio-variational linguistics and dialectology (eg. \cite{sali,labov,milroy1992linguistic,wolfram2005american}).
Since online content is an agglomeration of material originating from all over the world, language on the Internet demonstrates geographic variation.
The abundance of geo-tagged online text enables a study of geographic linguistic variation at scales that are unattainable using classical methods like surveys and questionnaires.

Characterizing and detecting such variation is challenging since it takes different forms: lexical, syntactic and semantic.  
Most existing work has focused on detecting lexical variation prevalent in geographic regions \cite{bamman2014gender,doyle2014mapping,eisenstein2010latent,eisenstein2014diffusion}.
However, regional linguistic variation is not limited to lexical variation. 

In this paper we address this gap. Our method, \dist, is the first computational approach for tracking and detecting statistically significant linguistic shifts of words across geographical regions. 
\dist\ detects syntactic and semantic variation in word usage across regions, in addition to purely lexical differences.
\dist\ builds on recently introduced neural language models that learn word representations (\emph{word embeddings}), extending them to capture region-specific semantics. 
Since observed regional variation could be due to chance, \dist\ explicitly introduces a null model to ensure detection of only statistically significant differences between regions.

Figure \ref{fig:cw} presents a visualization of the semantic variation captured by \dist\ for the word \texttt{test} between the United States, the United Kingdoms, Canada, and India. In the majority of English speaking countries, \texttt{test} almost always means an \texttt{exam}, but in India (where cricket is a popular sport) \texttt{test} almost always refers to a lengthy form of cricket match. One might argue that simple baseline methods like (analyzing part of speech) might be sufficient to identify regional variation. However because these methods capture different modalities, they detect different types of changes as we illustrate in Figure \ref{fig:crownjewel}.

We use our method in two novel ways.
First, we evaluate our methods on several large datasets at multiple geographic resolutions. 
We investigate linguistic variation across Twitter at multiple scales: (a) between four English speaking countries and (b) between fifty states in USA. 
We also investigate regional variation in the \ngrams\ data. Our methods detect a variety of changes including regional dialectical variations, region specific usages, words incorporated due to code mixing and differing semantics. 

\begin{figure}[ht!]
\vspace{-0.05in}
\begin{center}
\includegraphics[width=\columnwidth]{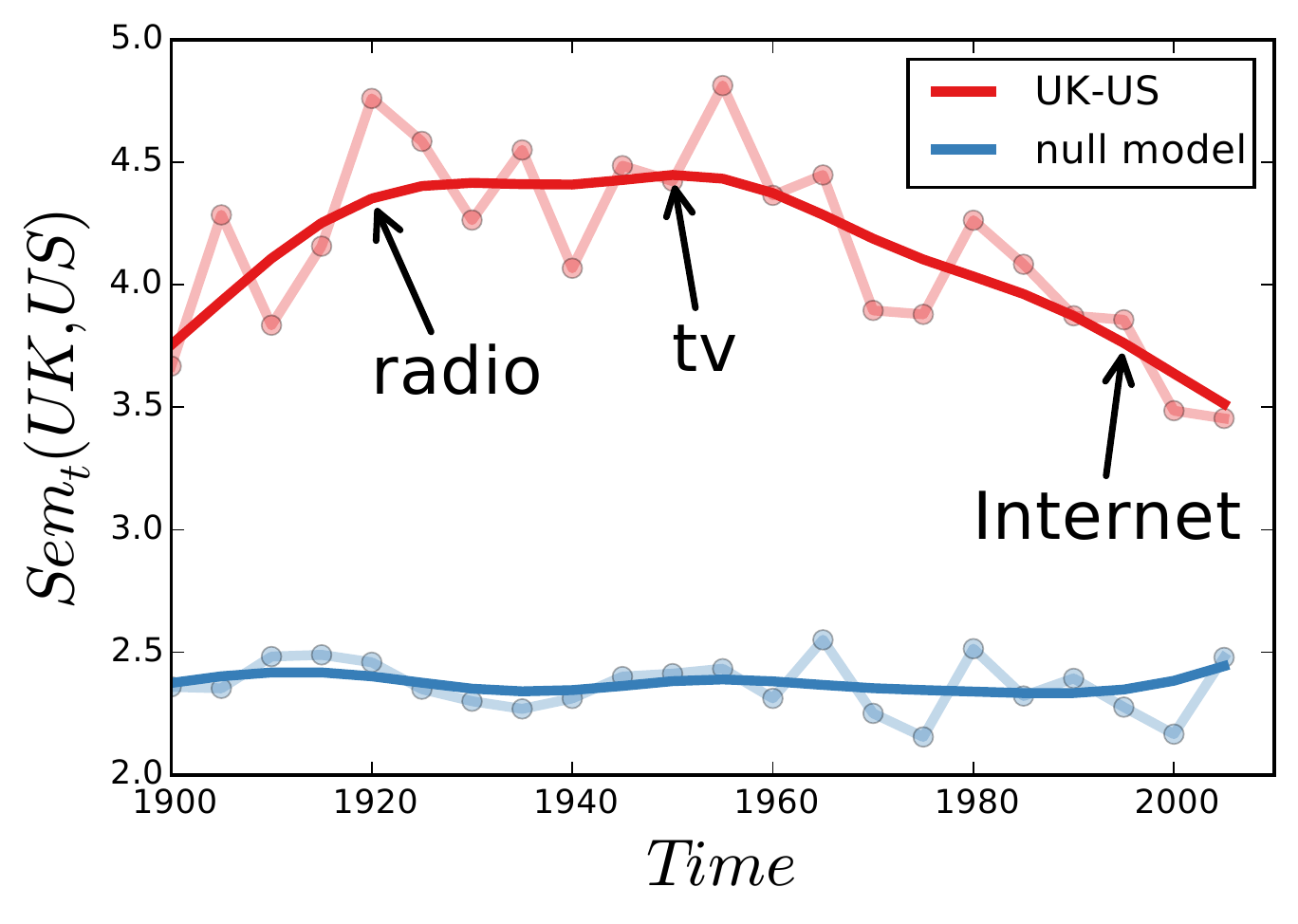}
\caption{Semantic Distance between UK English and US English at different time periods from 1900-2005. The two countries are becoming closer to one another driven by globalization and invention of mass communication technologies like radio, television, and the Internet.}
\label{fig:semantic_distance}
\end{center}
\vspace{-0.05in}
\end{figure}

Second, we apply our method to analyze distances between language dialects. 
In order to do this, we propose a measure of semantic distance between languages. 
Our analysis of British and American English over a period of $100$ years reveals that semantic variation between these dialects is shrinking potentially due to cultural mixing and globalization (see Figure \ref{fig:semantic_distance}). 

Specifically, our contributions are as follows:
\begin{itemize}
\item \textbf{Models and Methods}: 
We present our new method \dist\ which extends recently proposed neural language models to capture semantic differences between regions (Section \ref{sec:distributional}). 
\dist\ is a new statistical method that explicitly incorporates a null model to ascertain statistical significance of observed semantic changes.
\item \textbf{Multi-Resolution Analysis}: We apply our method on multiple domains (Books and Tweets) across geographic scales (States and Countries). 
Our analysis of these large corpora (containing billions of words) reveals interesting facets of language change at multiple scales of geographic resolution -- from neighboring states to distant continents (Section \ref{sec:results}).

\item \textbf{Semantic Distance}: We propose a new measure of semantic distance between languages which we use to characterize distances between various dialects of English and analyze their convergent and divergent patterns over time (Section \ref{sec:semanticdistance}).
\end{itemize}

\comment{
The rest of the paper is structured as follows:
In Section \ref{sec:def} we define the problem of linguistic variation over geography.
We then describe the various methods for capturing regional variation in word usage in Section \ref{sec:methods}.
In Section \ref{sec:changedetection}, we describe our method to ascertain statistical significance of changes.
We describe the datasets we used in Section \ref{sec:datasets}, and then comprehensively evaluate our methods in Section \ref{sec:results}. Our analysis of semantic distances between language dialects is discussed in Section \ref{sec:semanticdistance}.
We discuss related work in Section \ref{sec:related}, and conclude by discussing limitations and potential future work in Section \ref{sec:Conclusions}.
}

\section{Problem Definition}
We seek to quantify shift in word meaning (usage) across different geographic regions. 
Specifically, we are given a corpus $\mathcal{C}$ that spans $R$ regions where $\mathcal{C}_{r}$ corresponds to the corpus specific to region $r$. 
We denote the vocabulary of the corpus by $\mathcal{V}$. 
We want to detect words in $\mathcal{V}$ that have region specific semantics (not including trivial instances of words exclusively used in one region).
For each region $r$, we capture statistical properties of a word $w$'s usage in that region. 
Given a pair of regions $(r_{i}, r_{j})$, we then reduce the problem of detecting words that are used differently across these regions to an outlier detection problem using the statistical properties captured.

In summary, we answer the following questions:
\begin{enumerate}[noitemsep, topsep=1pt]
\item In which regions does the word usage drastically differ from other regions?
\item How statistically significant is the difference observed across regions?
\item Given two regions, how close are their corresponding dialects semantically?
\end{enumerate}  
\label{sec:def}
\section{Methods}
In this section we discuss methods to model regional word usage.
\label{sec:methods}
\subsection{Baseline Methods}
\label{sec:baselines}
\textbf{Frequency Method}.
\label{sec:freq}
\begin{figure}[t!]
\vspace{-0.05in}
  \centering
  \includegraphics[width=\columnwidth]{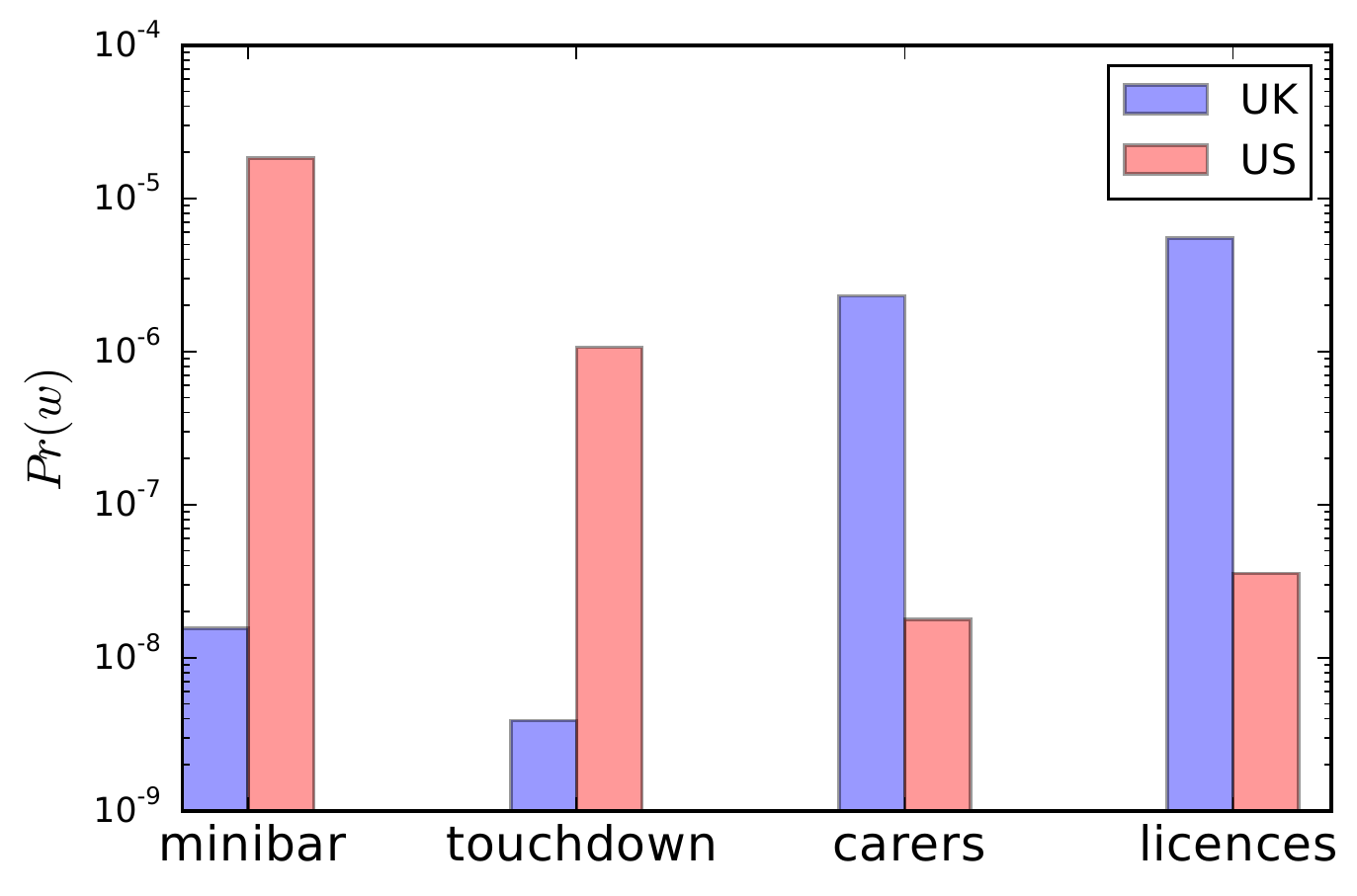}
\caption{Frequency usage of different words in English UK and English US. Note that \texttt{touchdown}, an American football term is much more frequent in the US than in UK. Words like \texttt{carers} and \texttt{licences} are used more in the UK than in the US. \texttt{carers} are known as \texttt{caregivers} in the US and \texttt{licences} is spelled as \texttt{licenses} in the US.}
\label{fig:freq}
\vspace{-0.05in}
\end{figure} 
One standard method to detect which words vary across geographical regions is to track their frequency of usage. 
Formally, we track the change in probability of a word across regions as described in \cite{kulkarni2015statistically}.
To characterize the difference in frequency usage of $w$ between a region pair $(r_{i}, r_{j})$, we compute the ratio $\textsc{Score}(w) = \frac{P_{r_{i}}(w)}{P_{r_{j}}(w)}$ where $P_{r_{i}}(w)$ is the probability of $w$ occurring in region $r_{i}$.
An example of the information we capture by tracking word frequencies over regions is shown in Figure \ref{fig:freq}.
Observe that \texttt{touchdown} (an American football term) is used much more frequently in the US than in UK.
While this naive method is easy to implement and identifies words which differ in their usage patterns, one limitation is an overemphasis on rare words. 
Furthermore frequency based methods overlook the fact that word usage or meaning changes are not exclusively associated with a change in frequency.

\textbf{Syntactic Method}.
\label{sec:syntactic}
A method to capture syntactic variation in word usage through time was proposed by \cite{kulkarni2015statistically}. Along similar lines, we can capture regional syntactic variation of words.
The word \texttt{lift} is a striking example of such variation: In the US, \texttt{lift} is dominantly used as a verb (in the sense: ``to lift an object''), whereas in the UK \texttt{lift} also refers to an elevator, thus predominantly used as a common noun.  
Given a word $w$ and a pair of regions $(r_{i},r_{j})$ we adapt the method outlined in \cite{kulkarni2015statistically} and compute the Jennsen-Shannon Divergence between the part of speech distributions for word $w$ corresponding to the regions.

\begin{figure}[tb!]
\vspace{-0.05in}
  \centering
  \includegraphics[width=\columnwidth]{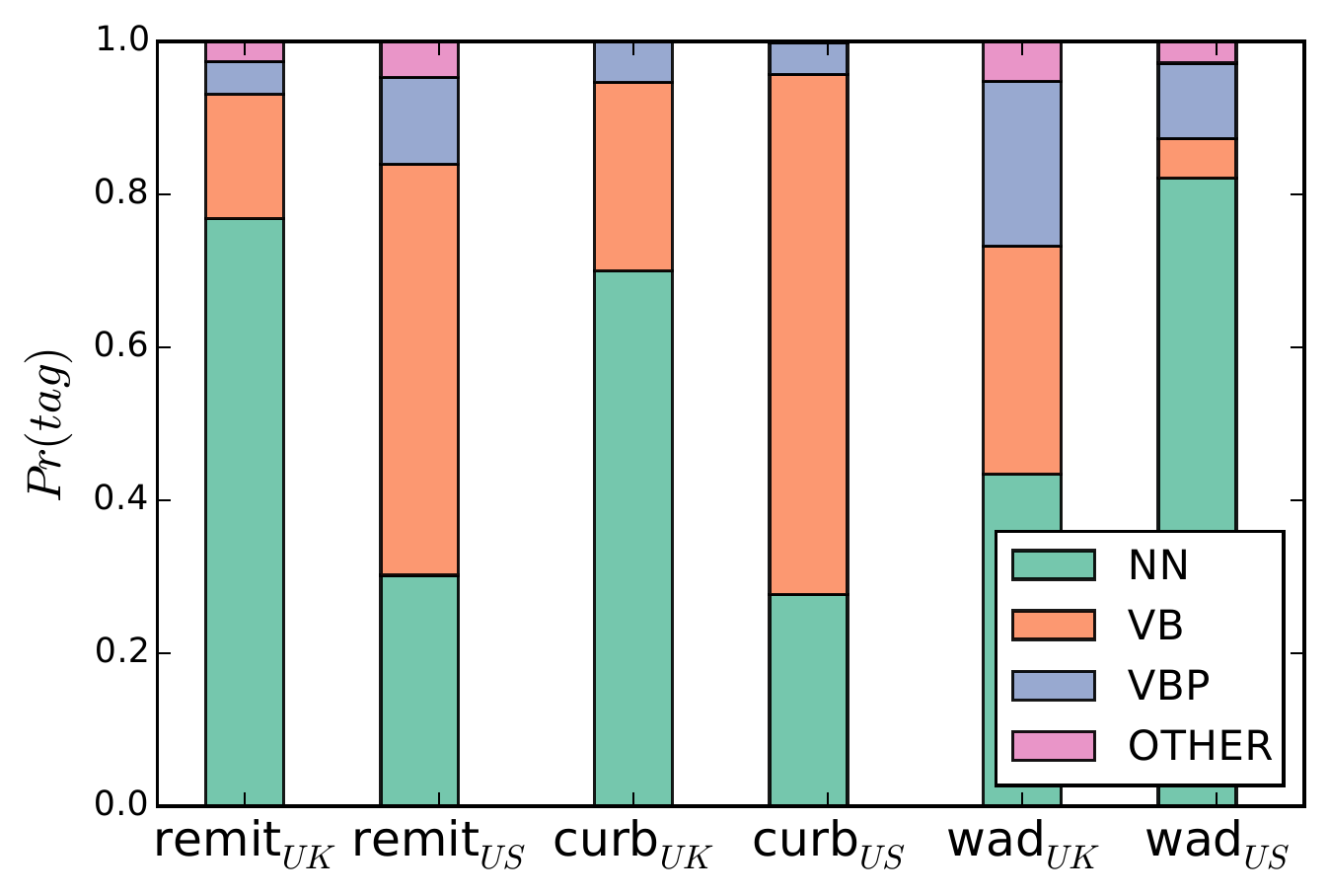}
\caption{Part of speech tag probability distribution of the words which differ in syntactic usage between UK and US. Observe that \texttt{remit} is predominantly  
used a verb (VB) in the US but as a common noun (NN) in the UK.}
\label{fig:pos}
\vspace{-0.05in}
\end{figure}

Figure \ref{fig:pos} shows the part of speech distribution for a few words that differ in syntactic usage between the US and UK.
In the US, \texttt{remit} is used primarily as a verb (as in ``to remit a payment'').
However in the UK, \texttt{remit} can refer ``to an area of activity over which a particular person or group has authority, control or influence'' (used as ``A remit to report on medical services'')\footnote{\url{http://www.oxfordlearnersdictionaries.com/us/definition/english/remit_1}}.
The word \texttt{curb} is used mostly as a noun (as ''I should put a curb on my drinking habits.'') in the UK but it is used dominantly as a verb in the US (as in ``We must curb the rebellion.'').  

Whereas the \syn\ method captures a deeper variation than the frequency methods, it is important to observe that semantic changes in word usage are not limited to syntactic variation as we illustrated before in Figure \ref{fig:crownjewel}.

\subsection{Distributional Method: \Large{\dist}}
\label{sec:distributional}
As we noted in the previous section, linguistic variation is not restricted only to syntactic variation. 
In order to detect subtle semantic changes, we need to infer cues based on the contextual usage of a word. 
To do so, we use distributional methods which learn a latent semantic space that maps each word $w \in \mathcal{V}$ to a continuous vector space $\mathbb{R}^{d}$.

We differentiate ourselves from the closest related work to our method \cite{bamman2014distributed}, by \emph{explicitly} accounting for random variation between regions, and proposing a method to detect statistically significant changes.

\subsubsection{Learning region specific word embeddings}  
Given a corpus $\mathcal{C}$ with $R$ regions, we seek to learn a region specific word embedding $\phi_{r}:\mathcal{V}, \mathcal{C}_r \mapsto \mathbb{R}^d$ using a neural language model. 
For each word $w\in\mathcal{V}$ the neural language model learns: 
\begin{enumerate} [noitemsep, topsep=1pt]
\item A global embedding $\delta_{\text{MAIN}}(w)$ for the word ignoring all region specific cues.
\item A differential embedding $\delta_{r}(w)$ that encodes differences from the global embedding specific to region $r$.
\end{enumerate}
The region specific embedding $\phi_{r}(w)$ is computed as: $\phi_{r}(w)=\delta_{\text{MAIN}}(w) + \delta_{r}(w)$. 
Before training, the global word embeddings are randomly initialized while the differential word embeddings are initialized to $\mathbf{0}$.
During each training step, the model is presented with a set of words $w$ and the region $r$ they are drawn from. 
Given a word $w_i$, the context words are the words appearing to the left or right of $w_i$ within a window of size $m$.
We define the set of active regions $\mathcal{A} = \{r, \text{MAIN}\}$ where $\text{MAIN}$ is a placeholder location corresponding to the global embedding and is always included in the set of active regions.
The training objective then is to maximize the probability of words appearing in the context of word $w_i$ conditioned on the active set of regions $\mathcal{A}$. Specifically, we model the probability of a context word $w_j$ given $w_i$ as:
\begin{align}
\Pr(w_j \mid w_i) &= \frac{\exp{(\mathbf{w}_j^T\mathbf{w}_i)}}
{\sum\limits_{w_k \in \mathcal{V}} \exp{(\mathbf{w}_k^T\mathbf{w}_i)}}
\label{eq:prob1}
\end{align} 
where $\mathbf{w}_i$ is defined as $\mathbf{w}_i = \sum\limits_{a \in \mathcal{A}} \delta_{a}(w_i)$.

During training, we iterate over each word occurrence in $\mathcal{C}$ to minimize the negative log-likelihood of the context words. 
Our objective function $J$ is thus given by:
\begin{align}
J &= \sum_{w_i \in \mathcal{C}} \sum_{\substack{j=i-m\\j!=i}}^{i+m} - \log \Pr(w_j \mid \mathbf{w}_i)
\label{eq:prob2}
\end{align}

When $|\mathcal{V}|$ is large, it is computationally expensive to compute the normalization factor in Equation \ref{eq:prob1} exactly. Therefore, we approximate this probability by using hierarchical soft-max \cite{hsm1,hsm2} which reduces the cost of computing the normalization factor from $\mathcal{O}(|\mathcal{V}|)$ to $\mathcal{O}(\log |\mathcal{V}|)$. 
We optimize the model parameters using stochastic gradient descent \cite{sgd}, as $\phi_t(w_i) = \phi_t(w_i) - \alpha \times \frac{\partial{J}}{\partial\phi_t(w_i)}$ where $\alpha$ is the learning rate.
We calculate the derivatives using the back-propagation algorithm \cite{backpropagation}.
We set $\alpha=0.025$, context window size $m$ to $10$ and size of the word embedding $d$ to be $200$ unless stated otherwise.

\begin{figure}[t!]
\vspace{-0.05in}
  \begin{center}
  \includegraphics[trim = 1in 0.7in 1in 0.7in, clip, width=\columnwidth]{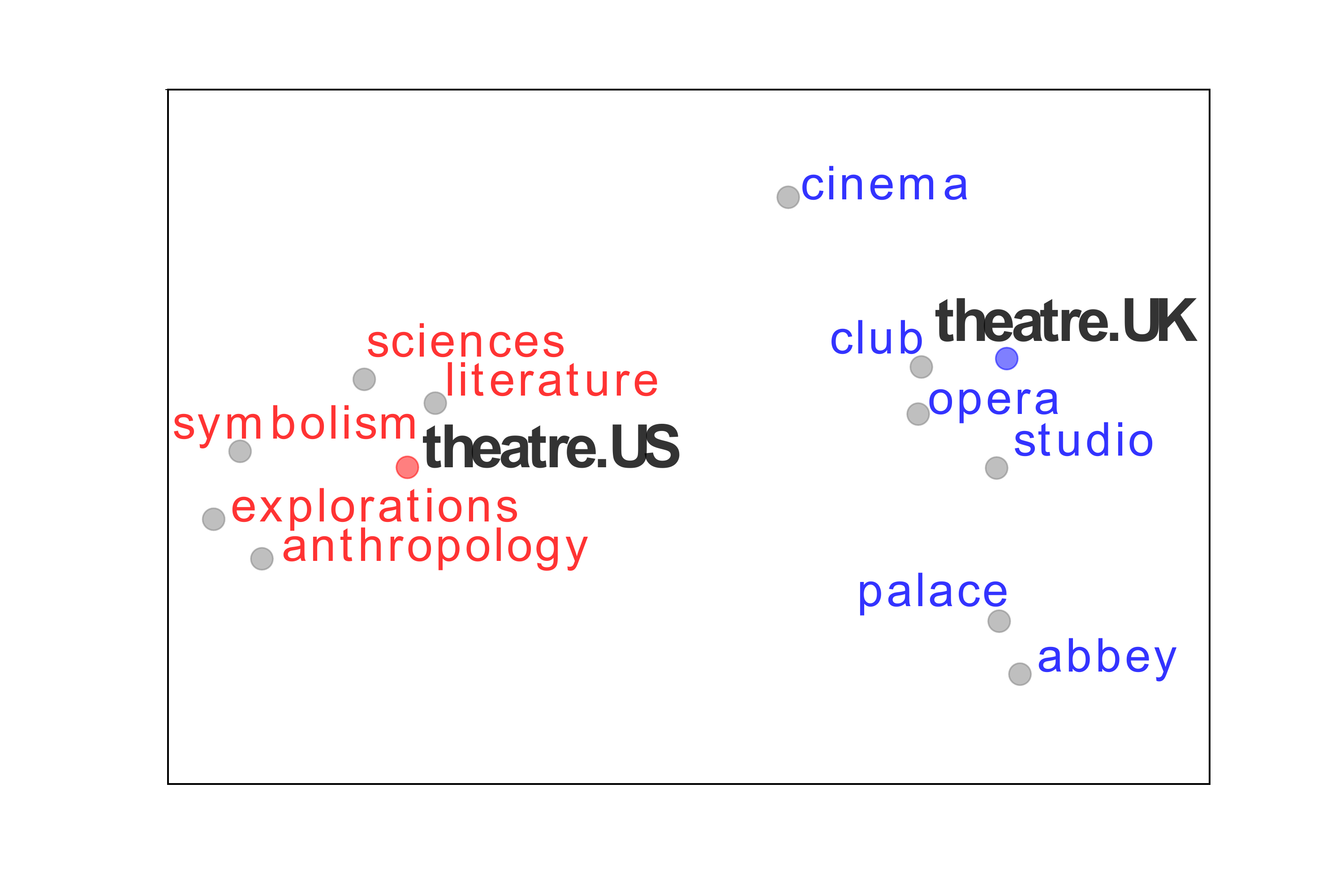}
\caption{Semantic field of \texttt{theatre} as captured by \dist\ method between the
UK and US. \texttt{theatre} is a field of study in the US while in the UK it primarily associated with opera or a club.}
  \label{fig:dist}
  \end{center}
\vspace{-0.1in}
\end{figure}

\subsubsection{Distance Computation between regional embeddings}
After learning word embeddings for each word $w\in{\mathcal{V}}$, we then compute the distance of a word between any two regions $(r_{i}, r_{j})$ as $\textsc{Score}(w) = \textsc{CosineDistance}(\phi_{r_{i}}(w), \phi_{r_{j}}(w))$
where $\textsc{CosineDistance}(u,v)$ is defined by $1 - \frac{u^{T}v}{\norm{u}_2\norm{v}_2}$. 

Figure \ref{fig:dist} illustrates the information captured by our \dist\ method as a two dimensional projection of the latent semantic space learned, for the word \texttt{theatre}.
In the US, the British spelling \texttt{theatre} is typically used only to refer to the performing arts.
Observe how the word \texttt{theatre} in the US is close to other subjects of study: \texttt{sciences, literature, anthropology}, but \texttt{theatre} as used in UK is close to places showcasing performances (like \texttt{opera}, \texttt{studio}, etc).
We emphasize that these regional differences detected by \dist\ are inherently \emph{semantic}, the result of a level of language understanding unattainable by methods which focus solely on lexical variation \cite{eisenstein2011discovering}.

\subsection{Statistical Significance of Changes}
\label{sec:changedetection}
In this section, we outline our method to quantify whether an observed change given by $\textsc{Score}(w)$ is significant. When one is operating on an entire population (or in the absence of stochastic processes), one fairly standard method to identify outliers is the $Z$-value test \cite{aggarwal2013outlier} (obtained by standardizing the raw scores) and marking samples whose $Z$-value exceeds a threshold $\beta$ (typically set to the $95$th percentile) as outliers. 

\begin{figure*}[tb]
        \vspace{-0.1in}
	\centering
	\begin{subfigure}{0.85\columnwidth}
		\includegraphics[width=\columnwidth]{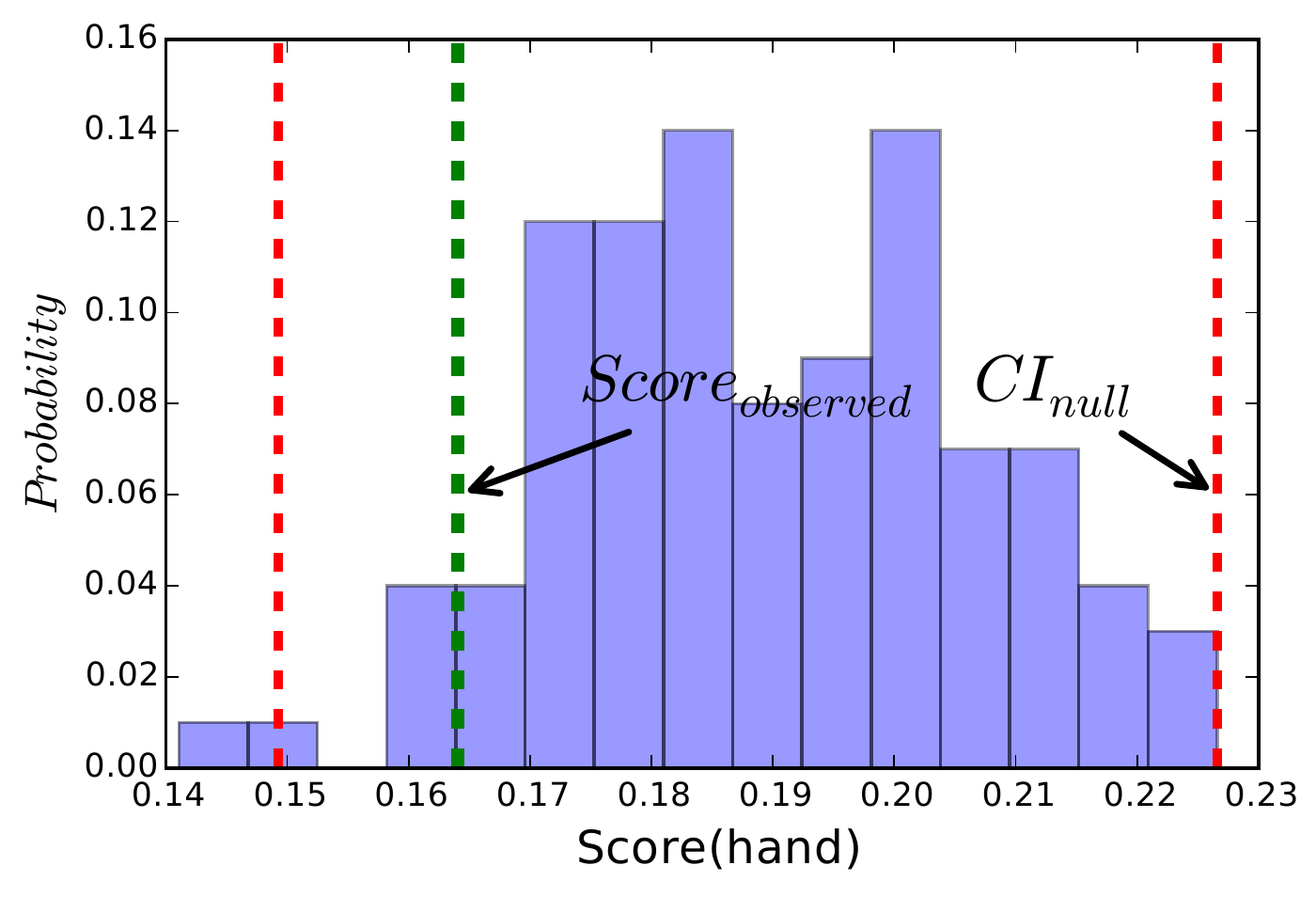}
		\caption{Observed score for \texttt{hand}}
		\label{fig:hand_null}
	\end{subfigure}
	\begin{subfigure}{0.85\columnwidth}
		\includegraphics[width=\columnwidth]{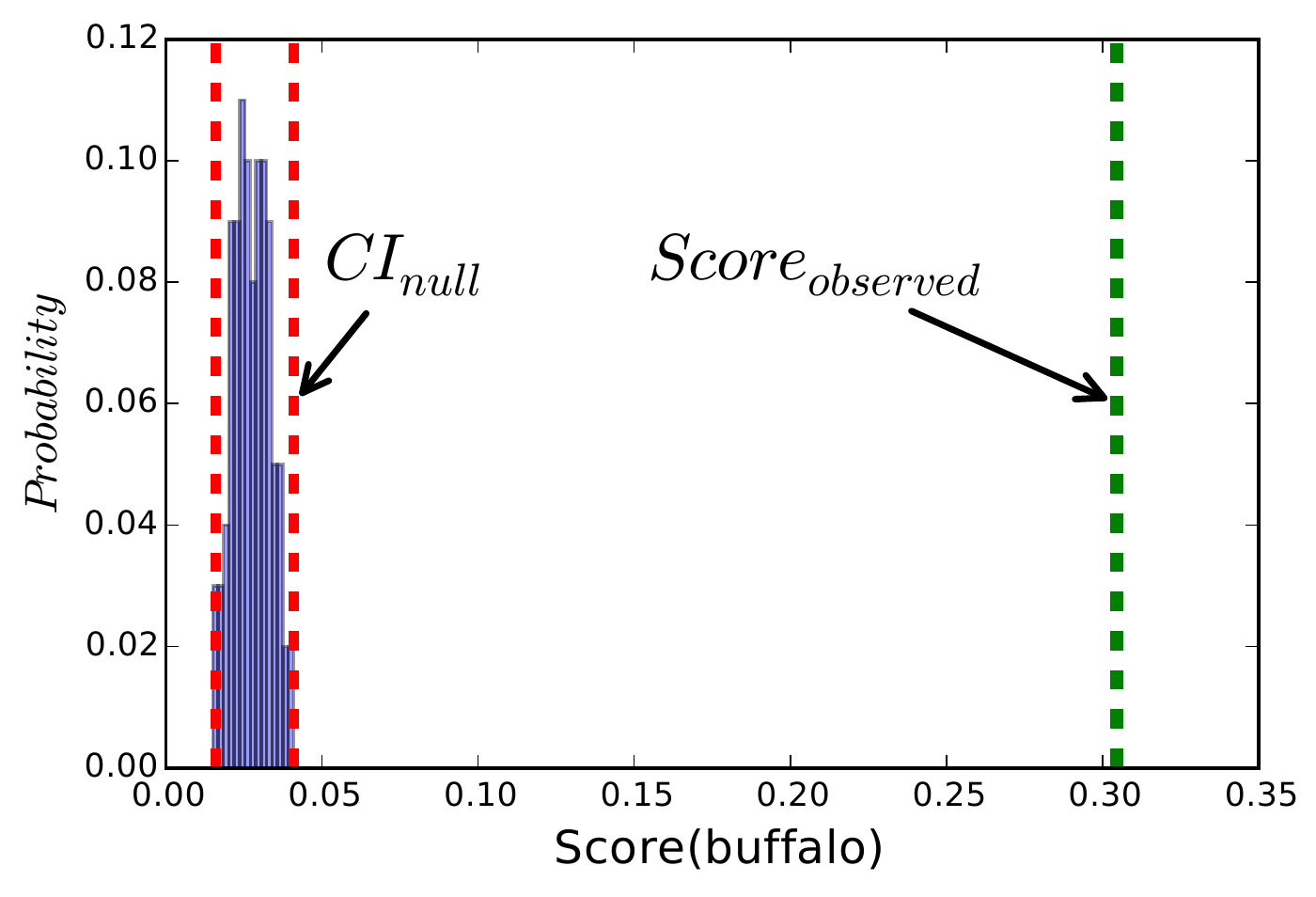}
		\caption{Observed score for \texttt{buffalo} }
		\label{fig:buffalo_null}
	\end{subfigure}
	\caption{The observed scores computed by \dist\ (in \protect \hwplotgreen) for \texttt{buffalo} and \texttt{hand} when analyzing regional differences between New York and USA overall. The histogram shows the distribution of scores under the null model. The $98\%$ confidence intervals of the score under null model are shown in \protect \hwplotred. The observed score for  \texttt{hand} lies well within the confidence interval and hence is not a statistically significant change. In contrast, the score for \texttt{buffalo} is far outside the confidence interval for the null distribution indicating a statistically significant change.} 
	\label{fig:change}
      \vspace{-0.1in}
\end{figure*}

However since in our method, $\textsc{Score}(w)$ could vary due random stochastic processes (even possibly pure chance), whether an observed score is significant or not depends on two factors: (a) the magnitude of the observed score (\emph{effect size}) and (b) probability of obtaining a score more extreme than the observed score, even in the absence of a true effect.

Specifically, given a word $w$ with a score $E(w)=\textsc{Score}(w)$ between regions $(r_{i}, r_{j})$ we ask the question: \emph{``What is the chance of observing $E(w)$ or a more extreme value assuming the absence of an effect?''}

First our method explicitly models the scenario when there is no effect, which we term as the \emph{null model}.
Next we characterize the distribution of scores under the null model. 
Our method then compares the observed score with this distribution of scores to ascertain the significance of the observed score. The details of our method are described in Algorithm \ref{alg:changedetect} and below. 

We simulate the null model by observing that under the null model, the labels of the text are \emph{exchangeable}. 
Therefore, we generate a corpus $C'$ by a random assignment of the labels (regions) of the given corpus $C$.
We then learn a model using $C'$ and estimate $\textsc{Score}(w)$ under this model. 
By repeating this procedure $B$ times we estimate the distribution of scores for each word under the null model (Lines \ref{lst:line:bsbegin} to \ref{lst:line:bsend}).

After we estimate the distribution of scores we then compute the $100\alpha\%$ confidence interval on $\textsc{Score}(w)$ under the null model. 
Thus for each word $w$, we specify two measures: (a) observed effect size and (b) $100\alpha\%$ confidence interval (we typically set $\alpha=0.95$) corresponding to the null distribution (Lines \ref{lst:line:hci}-\ref{lst:line:lci}). When the observed effect is not contained in the confidence interval obtained for the null distribution, the effect is statistically significant at the $1-\alpha$ significance level. 

Even though $p$-values have been traditionally used to report significance, recently researchers have argued against their use as $p$-values themselves do not indicate what the observed effect size was and hence even very small effects can be deemed statistically significant \cite{sullivan2012using, duconfidence}. In contrast, reporting effect sizes and confidence intervals enables us to factor in the magnitude of effect size while interpreting significance. In a nutshell therefore, we deem a change observed for $w$ as statistically significant when:
\begin{enumerate}[noitemsep, topsep=1pt]
	\item The effect size exceeds a threshold $\beta$ which ensures the effect size is large enough. One typically standardizes the effect size and typically sets $\beta$ to the $95$th percentile (which is usually around $3$).
	\item It is rare to observe this effect as a result of pure chance. This is captured by our comparison to the null model and the confidence intervals computed.
\end{enumerate}

Figure \ref{fig:change} illustrates this for two words: \texttt{hand} and \texttt{buffalo}. Observe that for \texttt{hand}, the observed score is smaller than the higher confidence interval, indicating that \texttt{hand} has not changed significantly. In contrast \texttt{buffalo} which is used differently in New York (since \texttt{buffalo} refers to a place in New York) has a score well above the higher confidence interval under the null model.

\begin{algorithm}[t!]
    \caption{\small \textsc{ScoreSignificance} ($C$, $B$, $\alpha$)}
    \label{alg:changedetect}
    \begin{algorithmic}[1]
        \REQUIRE $C$: Corpus of text with $R$ regions, $B$: Number of bootstrap samples, $\alpha$: Confidence Interval threshold
        \ENSURE $E$: Computed effect sizes for each word $w$, $\textsc{CI}$: Computed confidence intervals for each word $w$
        \STATEx // Estimate the NULL distribution.
        \STATE $\textsc{BS} \gets \emptyset $  \Comment {Corpora from the NULL Distribution}. $\textsc{NULLSCORES}(w)$ \Comment {Store the scores for $w$ under null model.} \label{lst:line:bsbegin}
        \REPEAT
                \STATE Permute the labels assigned to text of $C$ uniformly at random to obtain corpus $C'$
                \STATE $\textsc{BS} \gets \textsc{BS} \cup C'$
                \STATE Learn a model $N$ using $C'$ as the text.
                \FOR{$w \in \mathcal{V}$}
                     \STATE Compute $\textsc{Score}(w)$ using $N$.
                     \STATE Append $\textsc{Score}(w)$ to $\textsc{NULLSCORES}(w)$
                \ENDFOR 
        \UNTIL{$|\textsc{BS}|$ = $B$}
        \label{lst:line:bsend}
        \STATEx // Estimate the actual observed effect and compute confidence intervals.
        \STATE Learn a model $M$ using $C$ as the text.
        \FOR{$w \in \mathcal{V}$}
            \STATE  Compute $\textsc{Score}(w)$ using $M$.
            \STATE  $E(w) \gets \textsc{Score}(w)$
            \STATE  Sort the scores in $\textsc{NULLSCORES}(w)$. 
            \STATE  $\textsc{HCI}(w)$ $\gets$ $100\alpha$ percentile in $\textsc{NULLSCORES}(w)$
            \label{lst:line:hci}
            \STATE $\textsc{LCI}(w) \gets 100(1-\alpha)$ percentile in $\textsc{NULLSCORES}(w)$
            \label{lst:line:lci}
            \STATE $\textsc{CI}(w)$ $\gets$ $(\textsc{LCI}(w),\textsc{HCI}(w))$
        \ENDFOR 
        \STATE \textbf{return} $E, \textsc{CI}$  \label{lst:line:returnci}
    \end{algorithmic}
 \vspace{-0.01in}
\end{algorithm}

As we will also see in Section \ref{sec:results}, the incorporation of the null model and obtaining confidence estimates enables our method to efficaciously tease out effects arising due to random chance from statistically significant effects.
\vfill

\section{Datasets}
\label{sec:datasets}
Here we outline the details of two online datasets that we consider - Tweets from various geographic locations on Twitter and \ngrams.

\paragraph{The \ngrams} The \ngrams\ corpus \cite{Michel:ngramscorpus} contains frequencies of short phrases of text (\emph{ngrams}) which were taken from books spanning eight languages over five centuries. While these ngrams vary in size from $1-5$, we use the $5$-grams in our experiments. Specifically we use the \ngrams\ corpora for \emph{American English} and \emph{British English} and use a random sample of $30$ million ngrams for our experiments.
Here, we show a sample of 5-grams along with their region:
\begin{framed}
\texttt{\textbullet \; \small drive a coach and horses (\emph{UK}) \\
\indent
\textbullet \; \small years as a football coach (\emph{US}) 
}
\end{framed}
We obtained the POS Distribution of each word in the above corpora using Google Syntactic Ngrams\cite{syntacticngrams,syntactic2}.

\begin{table*}[tp!]
\vspace{-0.1in}
\begin{center}
		\begin{tabular}{l|l|S[table-format = <1.4]|p{90mm}}
			& \textbf{Word} & \textbf{US/UK $\Delta$} & \textbf{Explanation} \\
			\hline
			\parbox[t]{1em}{\multirow{3}{*}{\rotatebox[origin=c]{90}{\small \textbf{Books}}}}
			& \texttt{zucchini} & 2.3 & \emph{``zucchinis'' are known as ``courgettes'' in UK} \\
			& \texttt{touchdown} & 2.4 & \emph{``touchdown'' is a term in American football} \\
			& \texttt{bartender} & 2.5 & \emph{``bartender'' is a very recent addition to the pub language in UK.} \\			
			\multicolumn{4}{c}{} \\
			


			& \textbf{Word} & \textbf{US/UK $\Delta$} & \textbf{Explanation} \\			
			\hline			
			\parbox[t]{1em}{\multirow{7}{*}{\rotatebox[origin=c]{90}{\textbf{Tweets}}}}
			& \texttt{freshman} & 2.7 & \emph{``freshman'' are referred to as ``freshers'' in the UK} \\
			& \texttt{hmu} & 2.5 & \emph{hit me up a slang which is popular in USA} \\
			& & \textbf{US/AU $\Delta$} & \\			
			\cline{2-4}
			
			& \texttt{maccas} & -3.3 & \emph{McDonald's in Australia is called maccas} \\
			& \texttt{wickets} & -2.9 & \emph{wickets is a term in cricket, a popular game in Australia} \\
			& \texttt{heaps} & -2.7 & \emph{Australian colloquial for ``alot''} \\
			\hline
		\end{tabular}
	\caption{Examples of words detected by the \freq\ method on Google Book NGrams and Twitter. ($\Delta$ is difference in log probabilities between countries). A positive value indicates the word is more probable in the US than the other region. A negative value indicates the word is more probable in the other region than the US.}
	\label{tab:freq}
\end{center}	
\end{table*}

\begin{table*}[tp!]
\begin{center}
		\begin{tabular}{l|l|S[table-format = <1.4]|p{50mm}|p{60mm}}
			& \textbf{Word} & \textbf{JS} & \textbf{US Usage} & \textbf{UK Usage} \\
			\hline
			\parbox[t]{1em}{\multirow{3}{*}{\rotatebox[origin=c]{90}{\small \textbf{Books}}}} 
			& \texttt{remit} & 0.173  & \emph{remit the loan} & \emph{The jury investigated issues within its remit (an assigned area).} \\
			& \texttt{oracle} & 0.149  & \emph{Oracle the company} & \emph{a person who is omniscient} \\
			& \texttt{wad} & 0.143  & \emph{a wad of cotton} & \emph{Wad the paper towel and throw it! (used as ``to compress'')} \\
			
			\hline
			
			\parbox[t]{1em}{\multirow{10}{*}{\rotatebox[origin=c]{90}{\textbf{Tweets}}}}
			& \texttt{sort} & 0.224 & \emph{He's not a bad sort} & \emph{sort it out}   \\
			& \texttt{lift} & 0.220  & \emph{lift the bag} & \emph{I am stuck in the lift (elevator)} \\
			& \texttt{ring} & 0.200  & \emph{ring on my finger}  & \emph{give him a ring (call)}\\
			& \texttt{cracking} & 0.181 & \emph{The ice is cracking} & \emph{The girl is cracking (beautiful)}\\
			& \texttt{cuddle} & 0.148 & \emph{Let her cuddle the baby (verb)} &  \emph{Come here and give me a cuddle (noun)}\\
			& \texttt{dear} & 0.137 & \emph{dear relatives} & \emph{Something is dear (expensive)}\\

			& & & \textbf{US Usage} & \textbf{AU Usage} \\
			\cline{2-5}

			& \texttt{kisses} & 0.320 & \emph{hugs and kisses (as a noun)} & \emph{He kisses them (verb)}  \\
		   & \texttt{claim} & 0.109  & \emph{He made an insurance claim (noun)} & \emph{I claim ... (almost always used as a verb)} \\
		   \hline
		\end{tabular}
	\caption{Examples of words detected by the \syn\ method on Google Book NGrams and Twitter. (JS is Jennsen Shannon Divergence)}
	\label{tab:syn}
\end{center}
\end{table*}

\begin{table*}[tp!]
\begin{center}
\begin{subtable}[t]{\textwidth}
		\begin{tabular}{l|cc|p{45mm}|p{48mm}}
			
			\textbf{Word} & \textbf{Effect Size}  & \textbf{CI(Null)} & \textbf{US Usage} & \textbf{UK Usage} \\
			\hline
			
			\texttt{theatre} & 0.6067 & (0.004,0.007)  & \emph{great love for the theatre} & \emph{in a large theatre} \\
			
			\texttt{schedule} & 0.5153 & (0.032,0.050)  & \emph{back to your regular schedule} & \emph{a schedule to the agreement} \\
			
		    \texttt{forms}& 0.595 & (0.015, 0.026)   & \emph{out the application forms} & \emph{range of literary forms (styles)} \\
						
			\texttt{extract}& 0.400 & (0.023, 0.045)  & \emph{vanilla and almond extract}  & \emph{extract from a sermon} \\
			
			\texttt{leisure} & 0.535 & (0.012, 0.024) & \emph{culture and leisure (a topic)} & \emph{as a leisure activity}  \\
			
			\texttt{extensive} & 0.487 & (0.015, 0.027)& \emph{view our extensive catalog list}  & \emph{possessed an extensive knowledge (as in impressive)}  \\
			
			\texttt{store} & 0.423 & (0.02, 0.04)  & \emph{trips to the grocery store} & \emph{store of gold (used as a container)} \\
			
			\texttt{facility} & 0.378 & (0.035, 0.055)  & \emph{mental health,term care facility} & \emph{set up a manufacturing facility (a unit)} \\			
            \end{tabular}
            \caption{Google Book NGrams: Differences between English usage in the United States and United Kingdoms}
            \label{tab:ngrams_table_dist}
            \vspace{0.1in}
\end{subtable}

\begin{subtable}[b]{\textwidth}
		\begin{tabular}{l|cc|p{50mm}|p{48mm}}
		    \textbf{Word} & \textbf{Effect Size}  & \textbf{CI(Null)} & \textbf{US Usage} & \textbf{IN Usage} \\
					\hline
			\texttt{high} & 0.820 & (0.02,0.03) & \emph{I am in high school} & \emph{by pass the high way (as a road)} \\
			
			\texttt{hum} & 0.740 & (0.03, 0.04)  & \emph{more than hum and talk} & \emph{hum busy hain (Hinglish) } \\
			
			\texttt{main} & 0.691 & (0.048, 0.074) & \emph{your main attraction}  & \emph{main cool hoon (I am cool)} \\						
			\texttt{ring} & 0.718 & (0.054, 0.093) & \emph{My belly piercing ring} & \emph{on the ring road (a circular road)} \\
			
			\texttt{test} & 0.572 & (0.03, 0.061) & \emph{I failed the test} & \emph{We won the test} \\
			
			\texttt{stand} & 0.589 & (0.046, 0.07) & \emph{I can't stand stupid people} & \emph{Wait at the bus stand} \\			
		\end{tabular}		
	\caption{Twitter: Differences between English usage in the United States and India}
	\label{tab:ind_us_table_dist_twitter}
\end{subtable}
\caption{Examples of statistically significant geographic variation of language detected by our method, \dist, between English usage in the United States and English usage in the United Kingdoms (a) and India (b). ({CI} - the $98\%$ Confidence Intervals under the null model)}
\end{center}
\vspace{0.1in}
\end{table*}

\paragraph{Twitter Data} This dataset consists of a sample of Tweets spanning 24 months starting from September 2011 to October 2013.
Each Tweet includes the Tweet ID, Tweet and the geo-location if available.
We partition these tweets by their location in two ways:
\begin{enumerate}[noitemsep, topsep=1pt]
\item \emph{States in the USA}: We consider Tweets originating in the United States and group the Tweets by the state in the United States they originated from. The joint corpus consists of $7$ million Tweets. 
\item \emph{Countries}: We consider $11$ million Tweets originating from USA, UK, India (IN) and Australia (AU) and partition the Tweets among these four countries. 
\end{enumerate}
Some sample Tweet text is shown below:
\begin{framed}
	\texttt{\textbullet \; \small Someone come to golden with us! (\emph{CA}) \\
		\indent
		\textbullet \; \small Taking the subway with the kids ...(\emph{NY}) 
	}
\end{framed}

In order to obtain part of speech tags, for the tweets we use the TweetNLP POS Tagger\cite{owoputi2013improved}.


\section{Results and Analysis}
\label{sec:results}
In this section, we apply our methods to various data sets described above to identify words that are used differently across various geographic regions.  
We describe the results of our experiments below.

\comment{
\begin{figure}
\centering
\includegraphics[width=\columnwidth]{figs/synthetic_data_eval.pdf}
\caption{False Positive and False Negative Error Rates of our method as a function of effect size.}
\label{fig:synthetic_eval}
\end{figure}
}

\subsection{Geographical Variation Analysis}
Table \ref{tab:freq} shows words which are detected by the \freq\ method. Note that \texttt{zucchini} is used rarely in the UK because a \texttt{zucchini} is referred to as a \texttt{courgette} in the UK. Yet another example is the word \texttt{freshman} which refers to a student in their first year at college in the US. However in the UK a \texttt{freshman} is known as a \texttt{fresher}.
The \freq\ method also detects terms that are specific to regional cultures like \texttt{touchdown}, an American football term and hence used very frequently in the US. 

As we noted in Section \ref{sec:syntactic}, the \syn\ method detects words which differ in their syntactic roles.
Table \ref{tab:syn} shows words like \texttt{lift}, \texttt{cuddle} which are used as verbs in the US but predominantly as nouns in the UK. In particular \texttt{lift} in the UK also refers to an \emph{elevator}. 
While in the USA, the word \texttt{cracking} is typically used as a verb (as in ``the ice is cracking''), in the UK \texttt{cracking} is also used as an adjective and means ``stunningly beautiful''. The \freq\ method in contrast would not be able to detect such syntactic variation since it focuses only on usage counts and not on syntax.

In Tables \ref{tab:ngrams_table_dist} and \ref{tab:ind_us_table_dist_twitter} we show several words identified by our \dist\ method. While \texttt{theatre} refers primarily to a building (where events are held) in the UK, in the US \texttt{theatre} also refers primarily to the study of the performing arts. The word \texttt{extract} is yet another example: \texttt{extract} in the US refers to food extracts but is used primarily as a verb in the UK. While in the US, the word \texttt{test} almost always refers to an \texttt{exam}, in India \texttt{test} has an additional meaning of a cricket match that is typically played over five days. 
An example usage of this meaning is ``We are going to see the test match between India and Australia'' or the ``The test was drawn.''.
We reiterate here that the \dist\ method picks up on finer distributional cues that the \syn\ or the \freq\ method cannot detect. 
To illustrate this, observe that \texttt{theatre} is still used predominantly as a noun in both UK and the USA, but they differ in semantics which the \syn\ method fails to detect. 

\begin{table*}[ht!]
\centering
\begin{tabular}{m{2cm}|m{4cm} m{4cm} m{4cm}}
\textbf{Word} & \multicolumn{3}{c}{\textbf{Distances}} \\
 & & & \\
 & Naive Distances & \textsc{nullmodel} & \dist (\textbf{Our Method}) \\ \hline
 & & &  \\
\texttt{buffalo} & \includegraphics[scale=0.25]{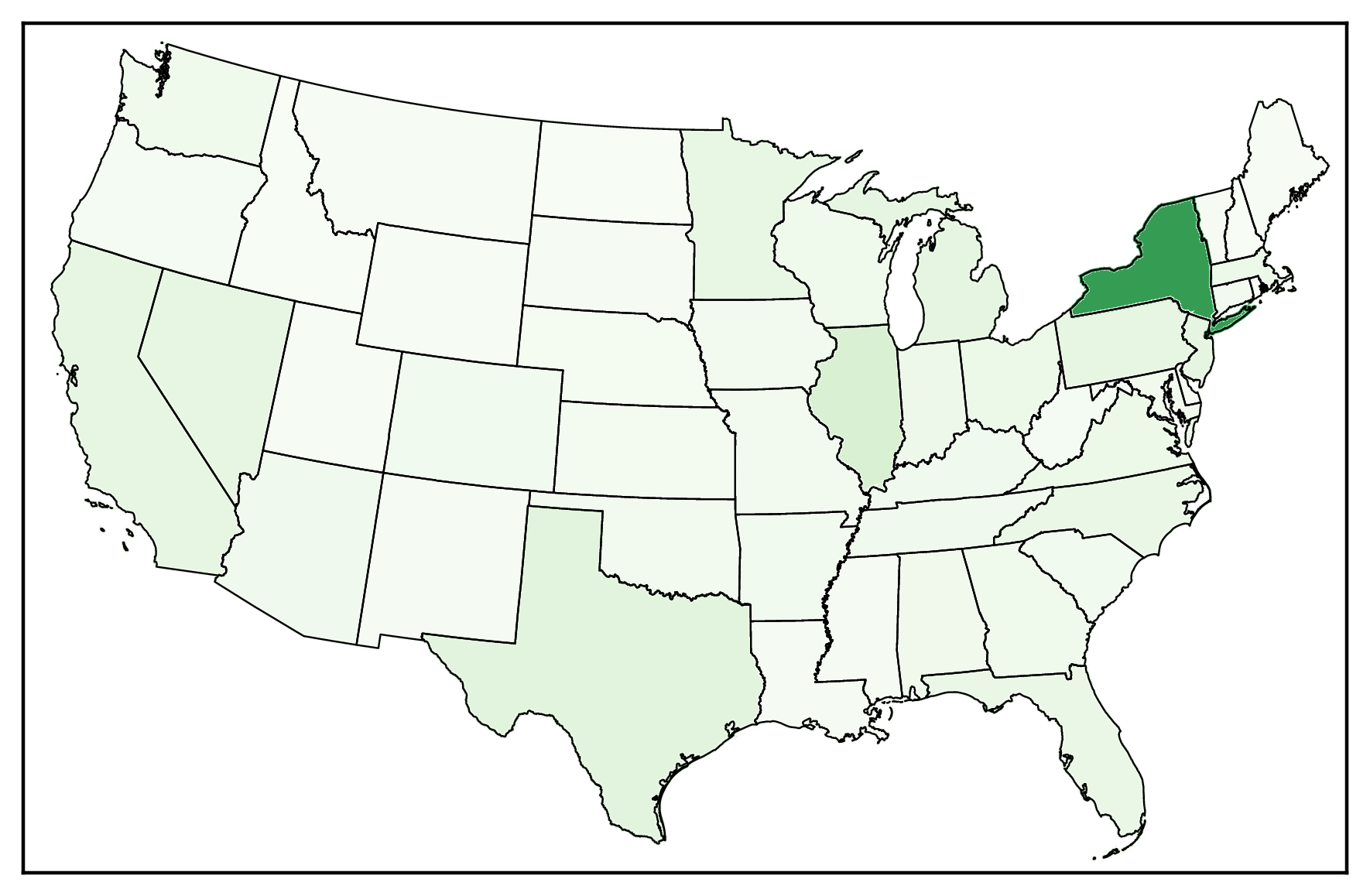} & \includegraphics[scale=0.25]{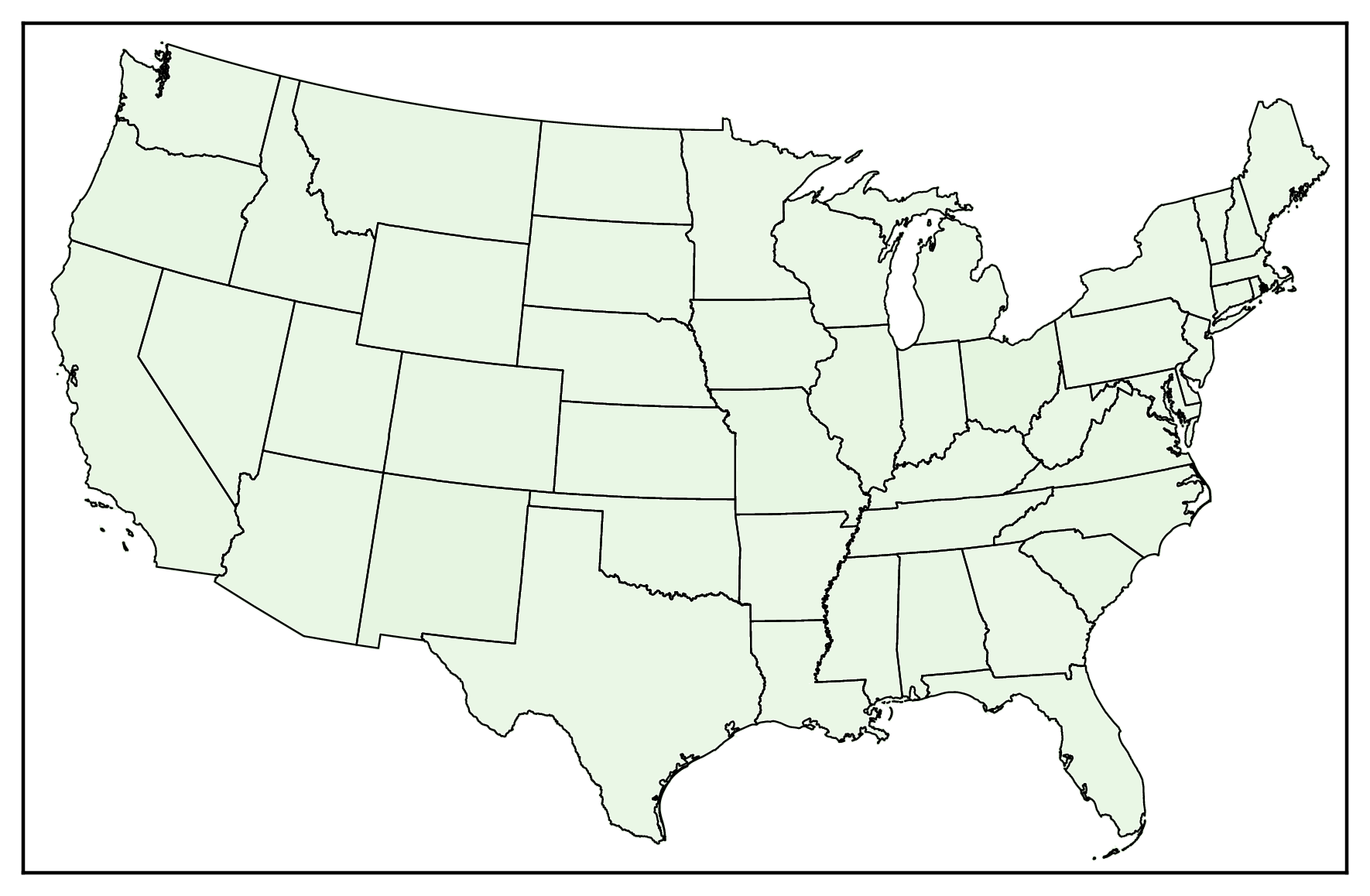}  & \includegraphics[scale=0.29]{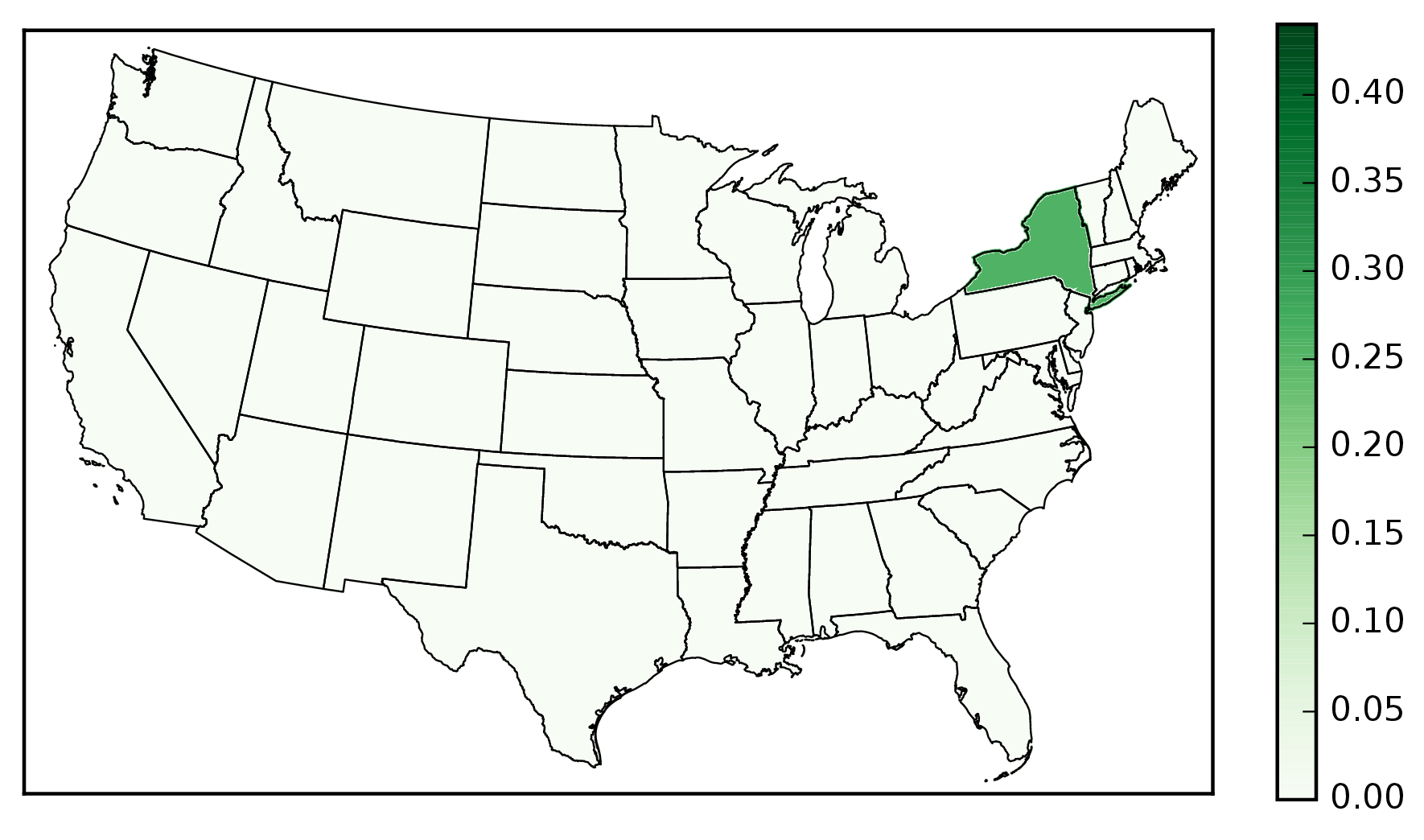} \\
\texttt{twins} & \includegraphics[scale=0.25]{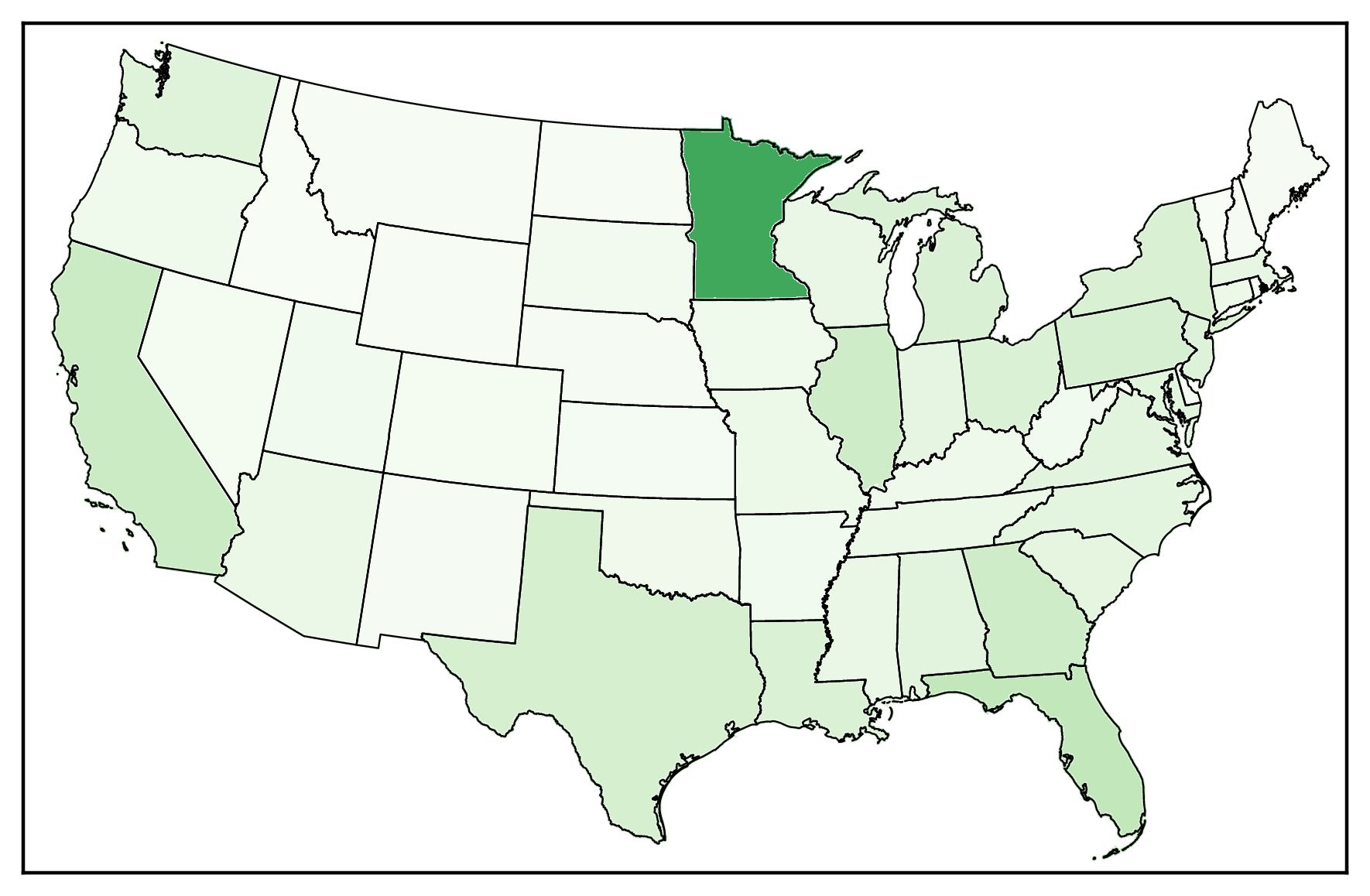} & \includegraphics[scale=0.25]{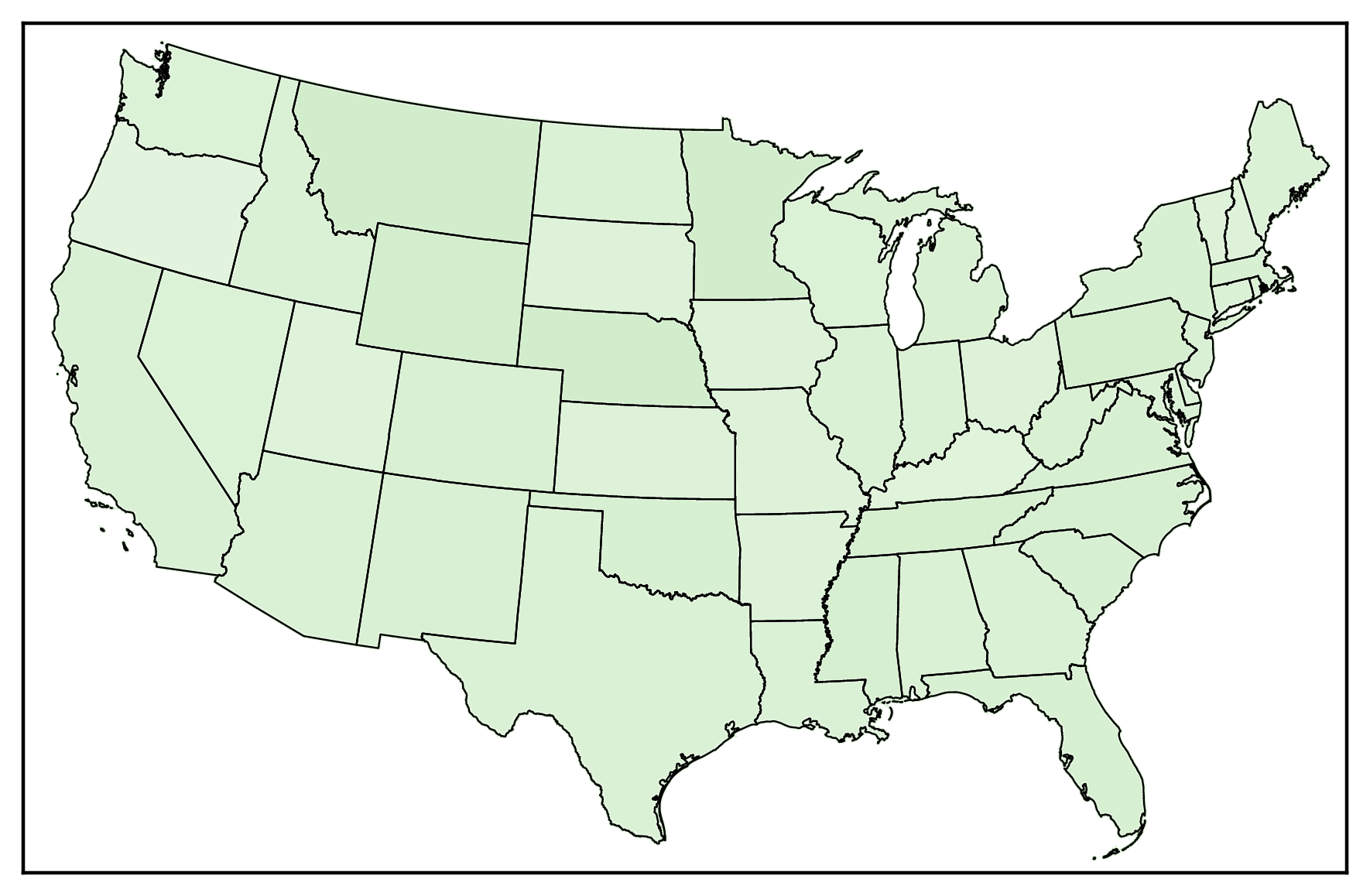}  & \includegraphics[scale=0.29]{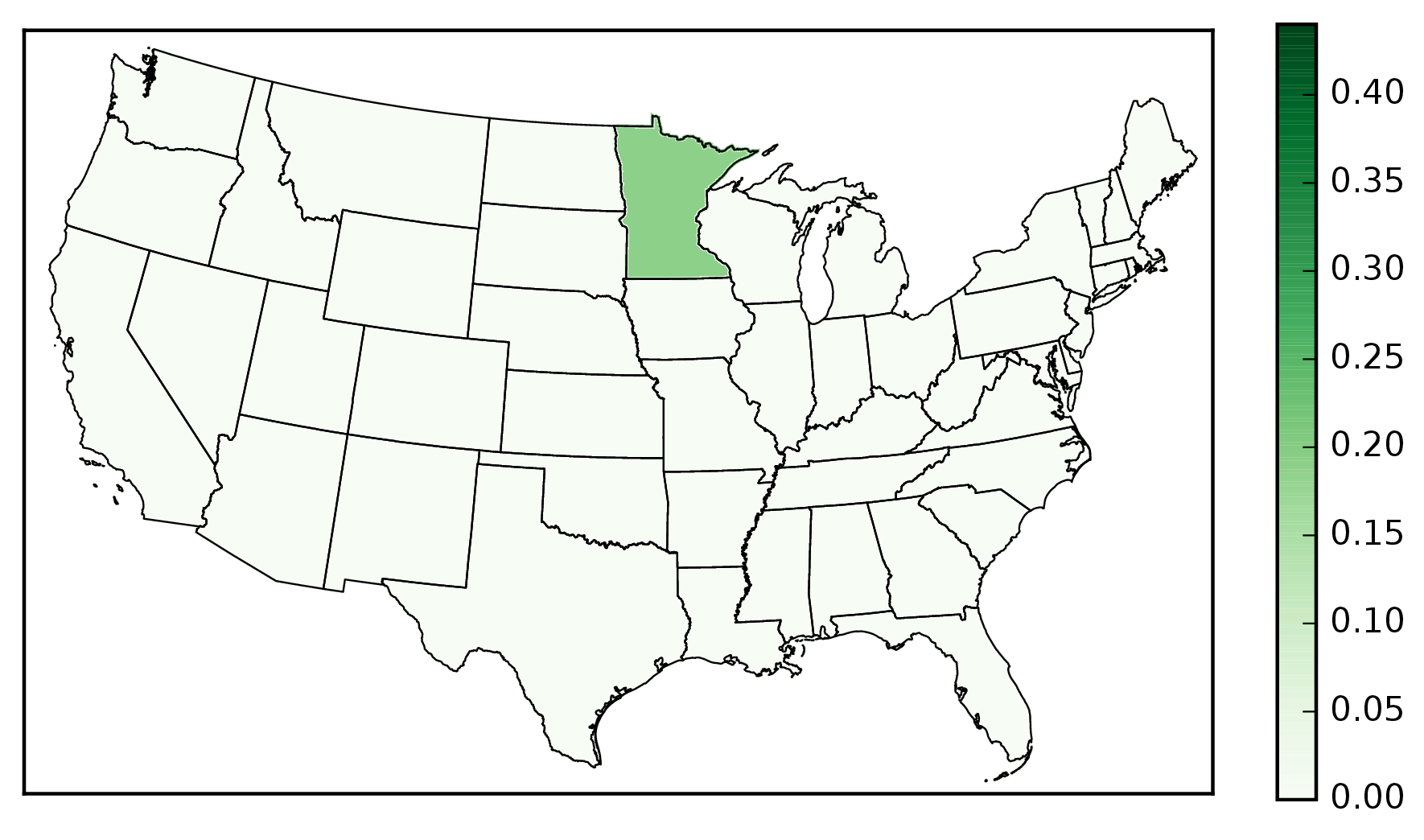} \\
\texttt{space} & \includegraphics[scale=0.25]{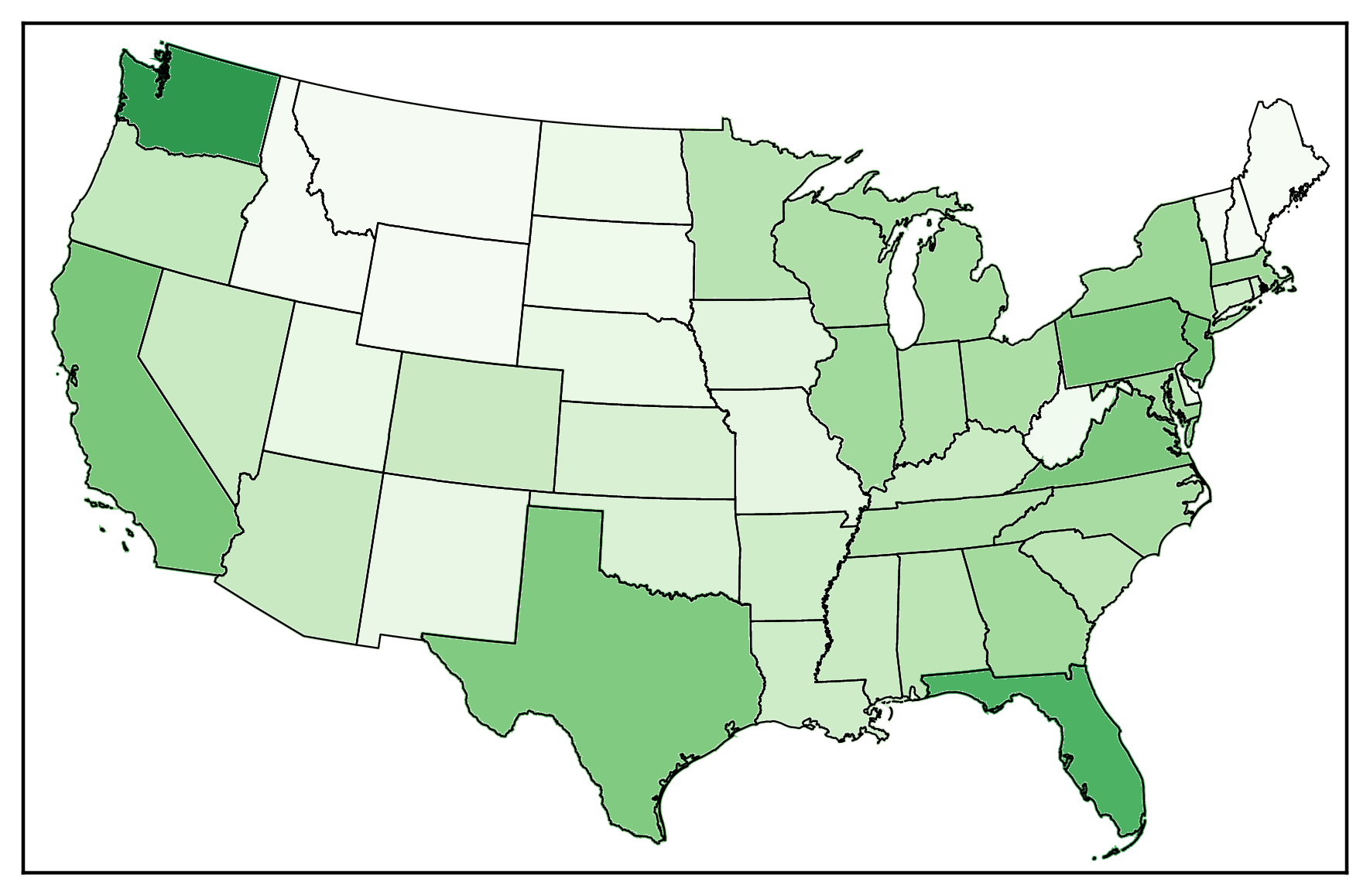} & \includegraphics[scale=0.25]{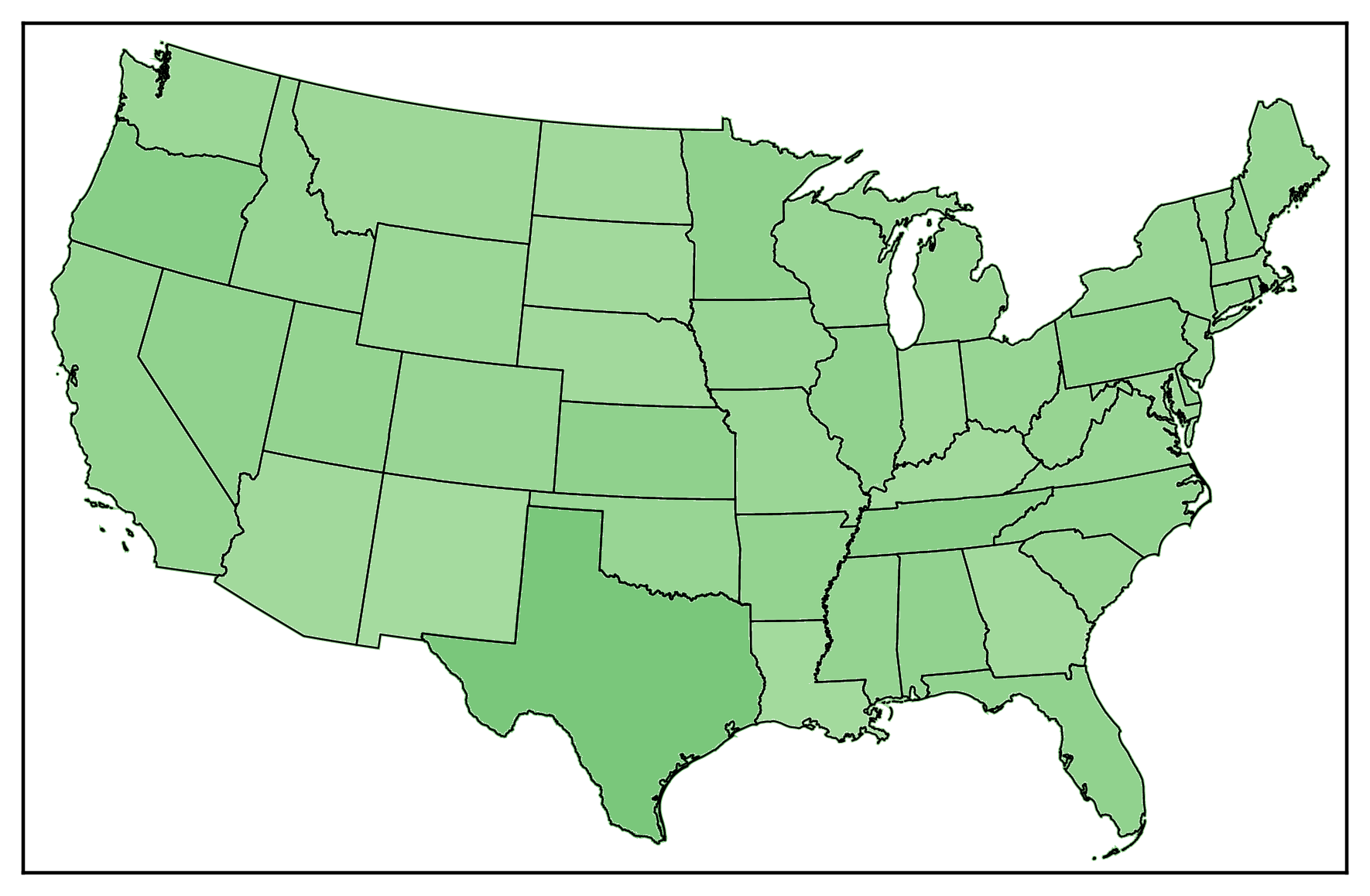}  & \includegraphics[scale=0.29]{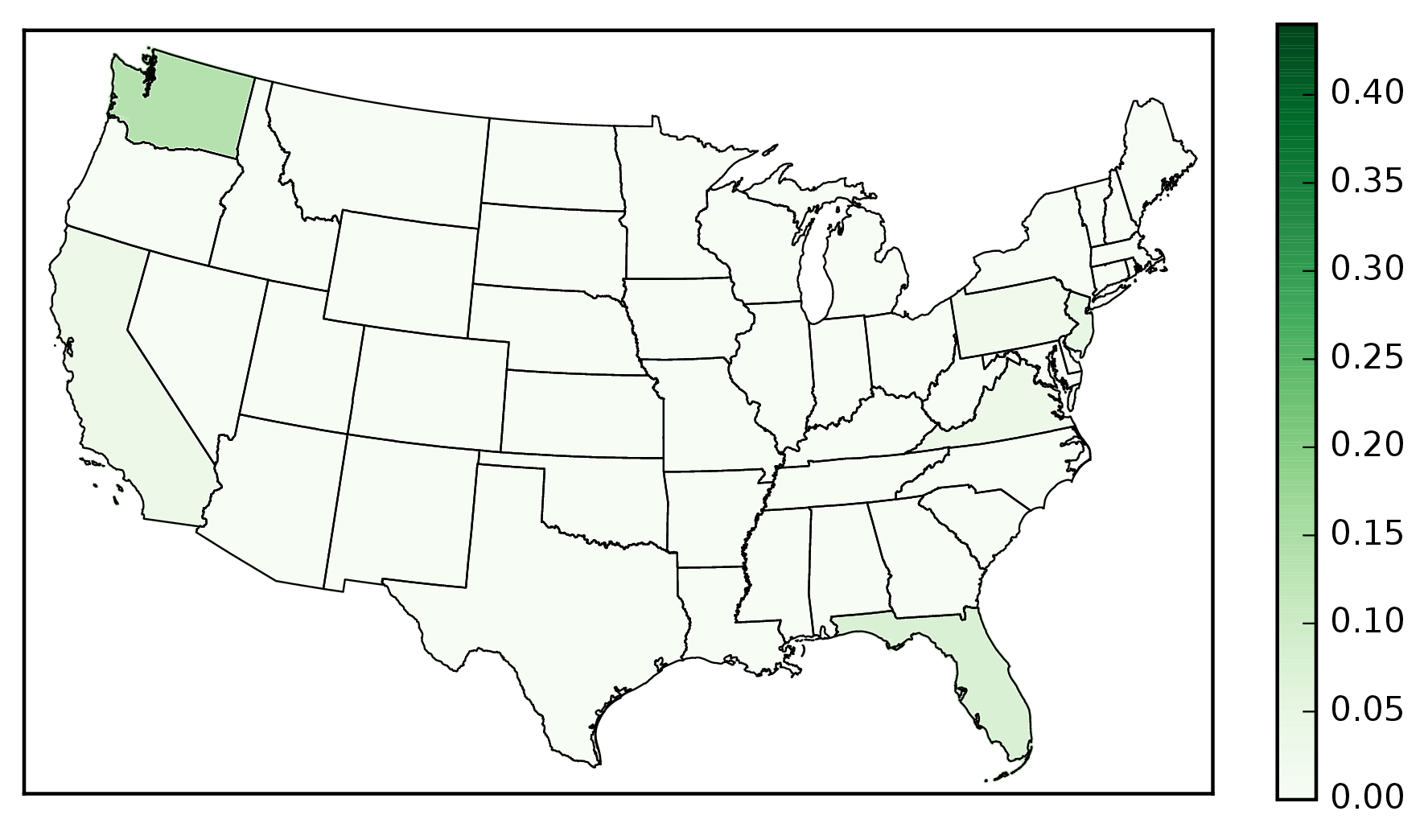} \\
\texttt{golden} & \includegraphics[scale=0.25]{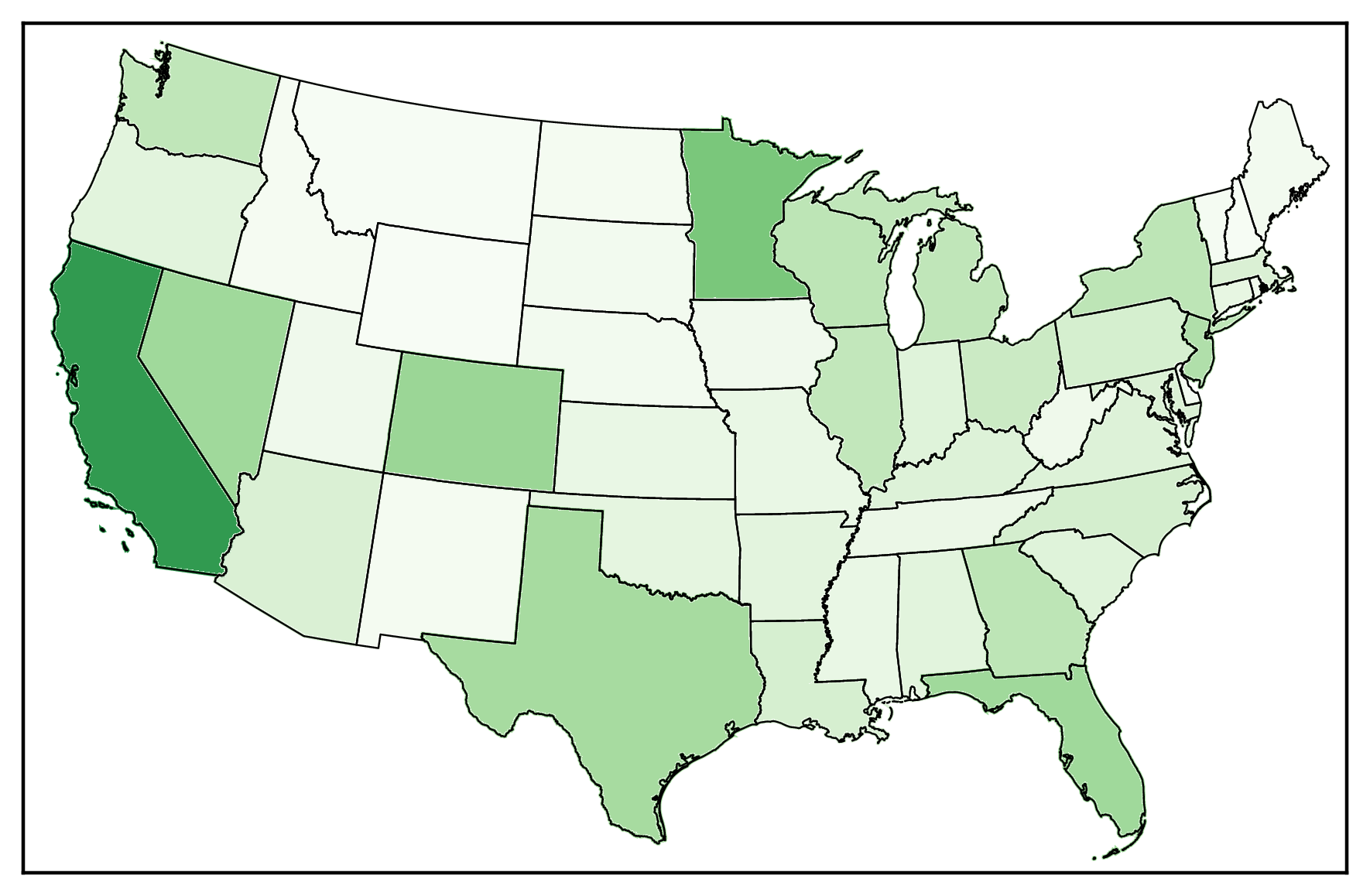} & \includegraphics[scale=0.25]{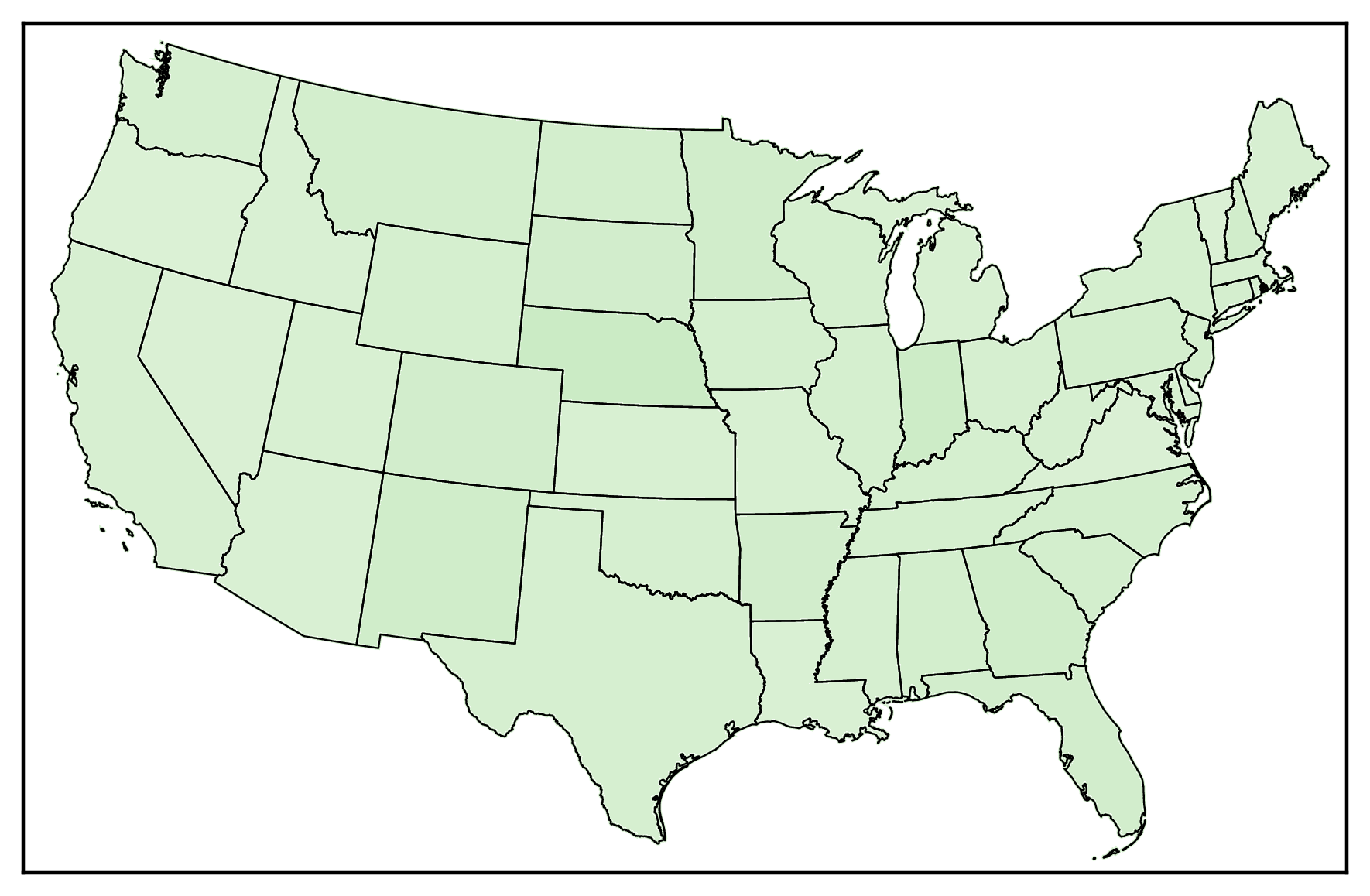}  & \includegraphics[scale=0.29]{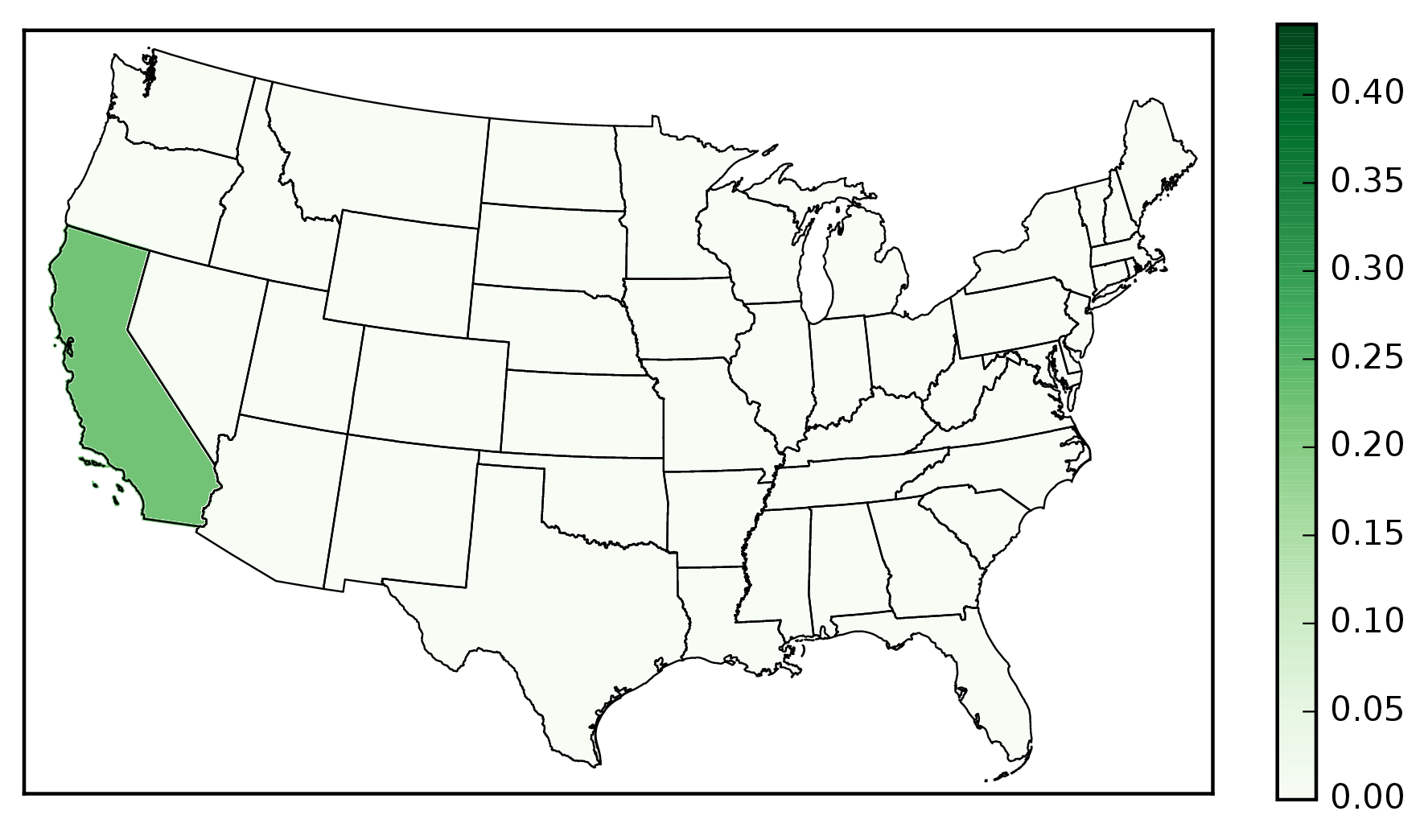} \\
\texttt{hand} & \includegraphics[scale=0.25]{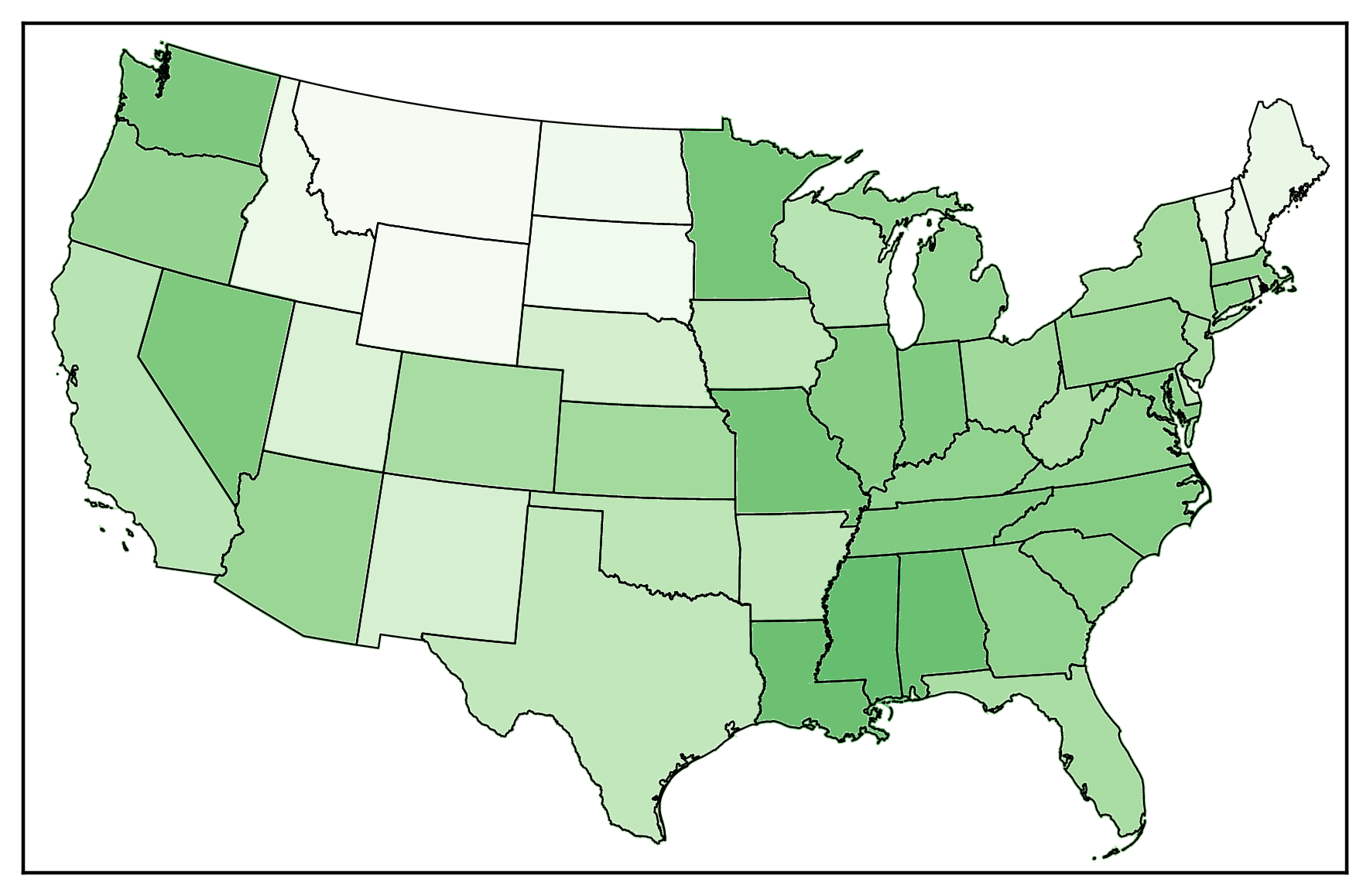} & \includegraphics[scale=0.25]{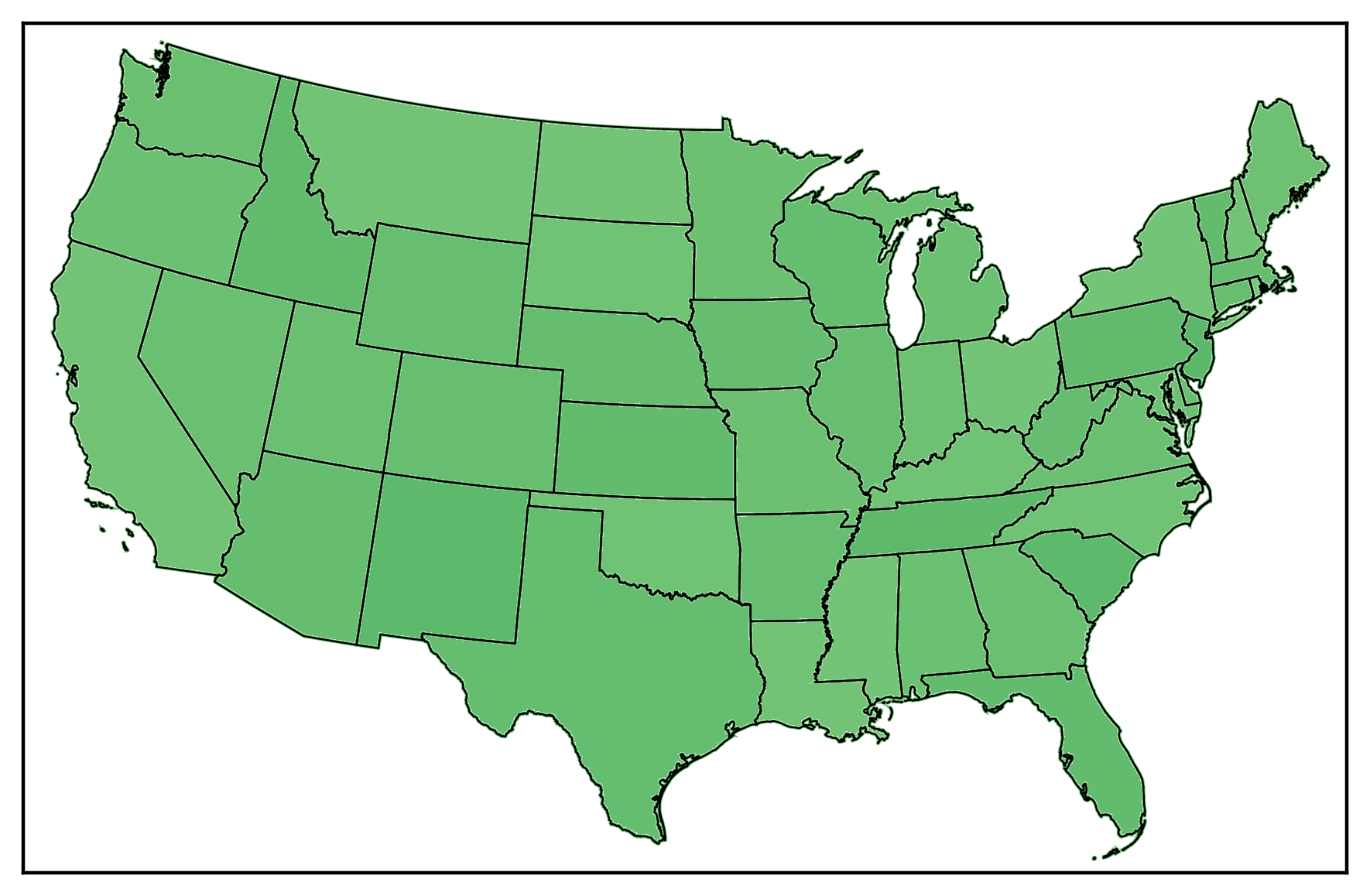}  & \includegraphics[scale=0.29]{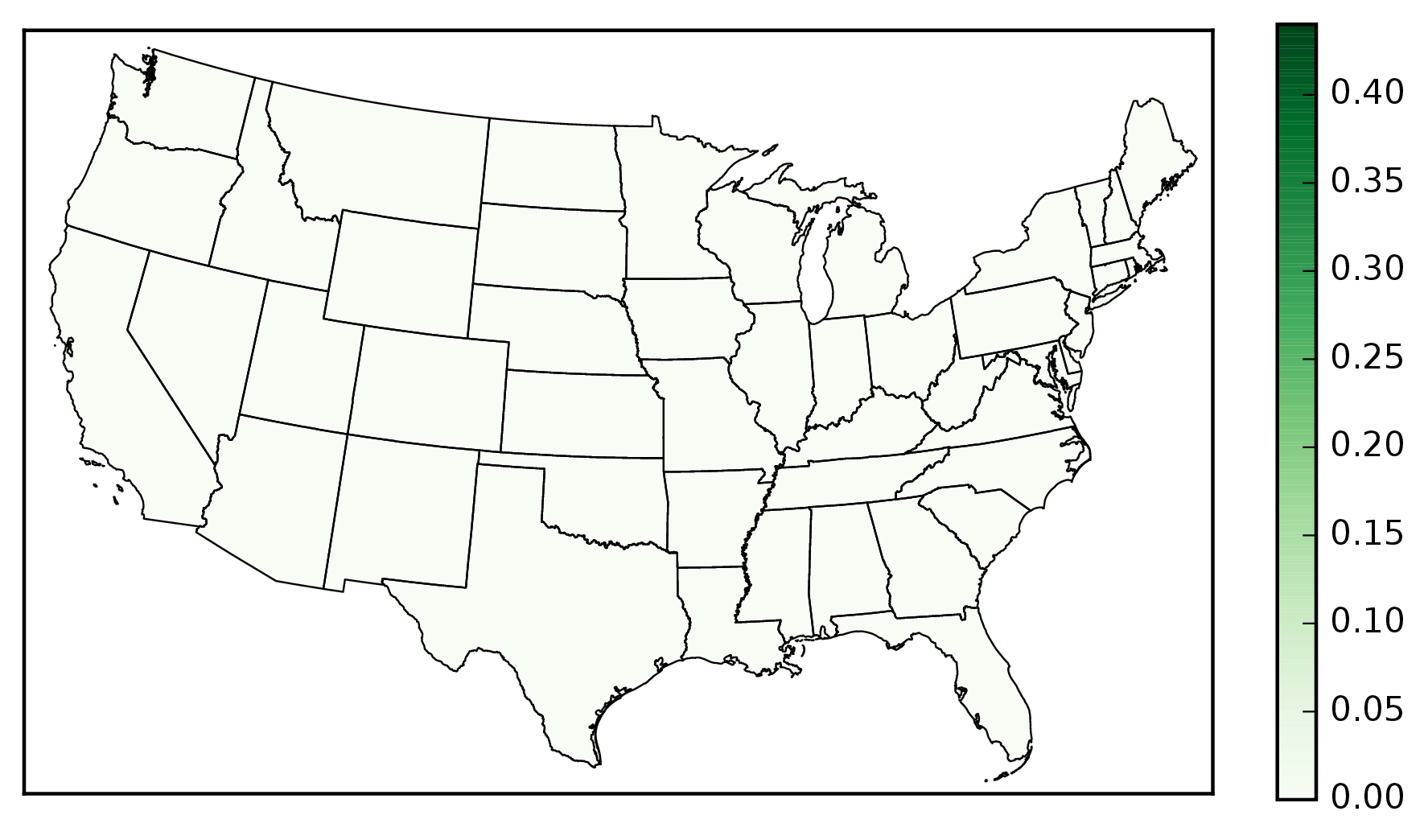} \\
\end{tabular}
\caption{Sample set of words which differ in meaning (semantics) in different states of the USA. Note how incorporating the null model highlights only statistically significant changes. Observe how our method \dist\ correctly detects no change in \texttt{hand}.}
\label{tab:gt}
\end{table*}

Another clear pattern that emerges are ``code-mixed words'', which are regional language words that are incorporated into the variant of English (yet still retaining the meaning in the regional language).
Examples of such words include \texttt{main} and \texttt{hum} which in India also mean ``I'' and ``We'' respectively in addition to their standard meanings.  In Indian English, one can use \texttt{main} as ``the main job is done'' as well as ``main free at noon. what about you?''. 
In the second sentence \texttt{main} refers to ``I'' and means ``I am free at noon. what about you?''.

Furthermore, we demonstrate that our method is capable of detecting changes in word meaning (usage) at finer scales (within states in a country).
Table \ref{tab:gt} shows a sample of the words in states of the USA which differ in semantic usage markedly from their overall semantics globally across the country. 

Note that the usage of \texttt{buffalo} significantly differs in New York as compared to the rest of the USA. \texttt{buffalo }typically would refer to an animal in the rest of USA, but it refers to a place named \emph{Buffalo} in New York. The word \texttt{queens} is yet another example where people in New York almost always refer to it as a place.

Other clear trends evident are words that are typically associated with states.
Examples of such words include \texttt{golden}, \texttt{space} and \texttt{twins}. The word \texttt{golden} in California almost always refers to \emph{The golden gate bridge} and \texttt{space} in Washington refers to \emph{The space needle}. While \texttt{twins} in the rest of the country is dominantly associated with twin babies (or twin brothers), in the state of Minnesota, \texttt{twins} also refers to the state's baseball team \emph{Minnesota Twins}.

Table \ref{tab:gt} also illustrates the significance of incorporating the null model to detect which changes are significant. 
Observe how incorporating the null model renders several observed changes as being not significant thus highlighting statistically significant changes. 
Without incorporating the null model, one would erroneously conclude that \texttt{hand} has different semantic usage in several states. However on incorporating the null model, we notice that these are very likely due to random chance thus enabling us to reject this as signifying a true change.

These examples demonstrate the capability of our method to detect wide variety of variation across different scales of geography spanning regional differences to code-mixed words.

\section{Semantic Distance}
\label{sec:semanticdistance}
\begin{figure}[t!]
		\begin{center}
		\includegraphics[width=0.75\columnwidth]{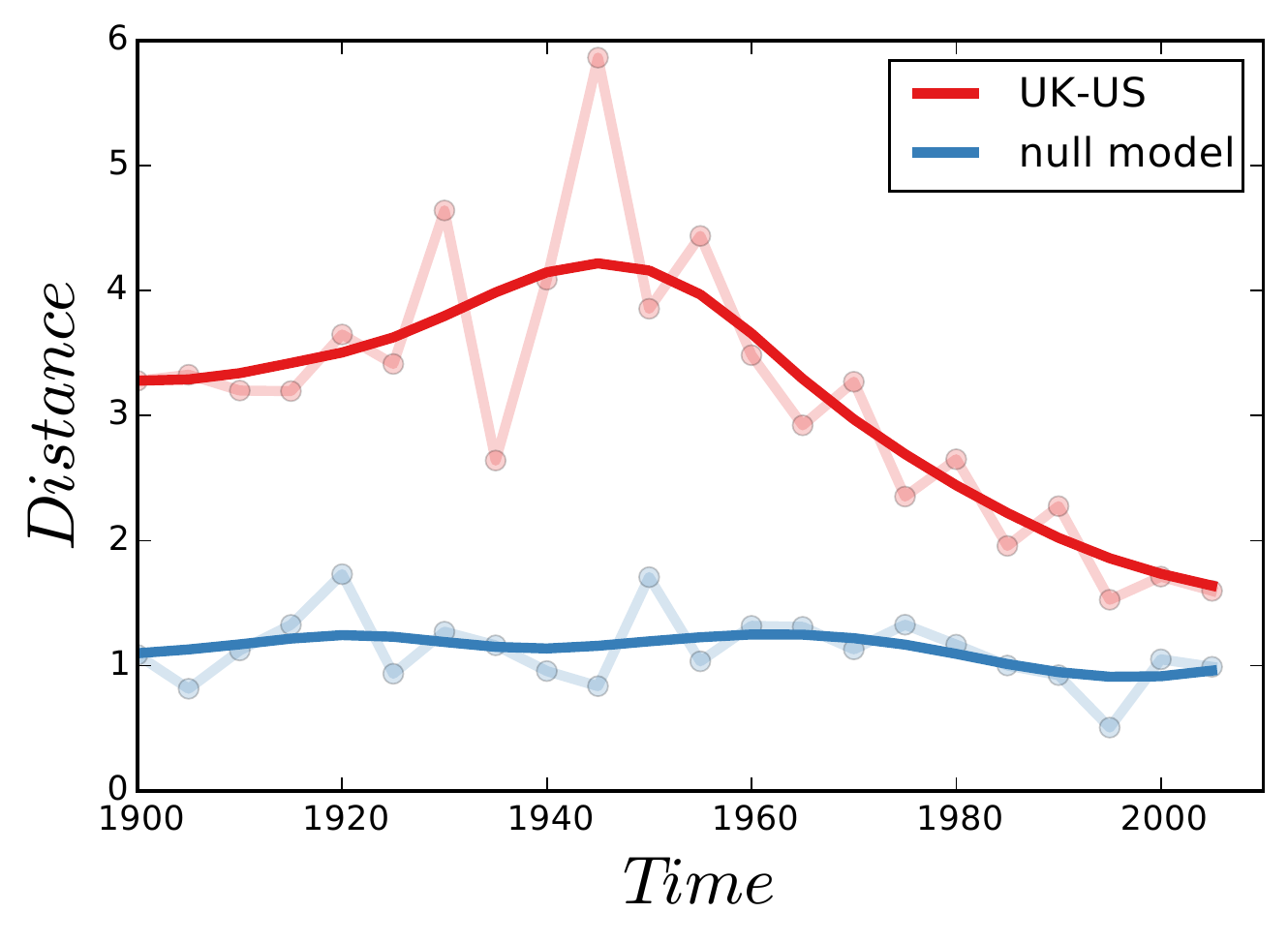}
		\caption{Usage of \texttt{acts} in UK converges to the usage in US over time.}
		\label{fig:acts}	
		\end{center}
\end{figure}
		
In this section we investigate the following question: Are British and American English converging or diverging over time semantically?

In order to measure semantic distance between languages through time, we propose a measure of semantic distance between two variants of the language at a given point $t$. Specifically, at a given time $t$, we are given a corpus $\mathcal{C}$ and a pair of regions $(r_{i},r_{j})$. Using our method (see Section \ref{sec:distributional}) we compute the standardized distance $Z_{t}(w)$ for each word $w$ between the regions at time point $t$. 
Then, we construct the intersection of the set of words $\mathcal{W}$ that have been deemed to have changed significantly at each time point $t$. 
We do this so that (a) we focus on only the words that were significantly different between the language dialects at time point $t$ and (b) the words identified as different are stable across time, allowing us to track the usage of the same set of divergent words over time. Our measure of the semantic distance between the two language dialects at time $t$ is then $ Sem_{t}(r_{i}, r_{j}) =\frac{1}{|\mathcal{W}|} \sum_{w \in \mathcal{W}}Z_{t}(w)$, the mean of the distances of words in $\mathcal{W}$.

In our experiment, we considered the \ngrams\ for UK English and US English within a time span of $1900-2005$ using a window of $5$ years. We computed the semantic distance between these dialects as described above, which we present in Figure \ref{fig:semantic_distance}. We clearly observe a trend showing both British and American English are converging. Figure \ref{fig:acts} shows one such word \texttt{acts}, where the usage in the UK starts converging to the usage in the US. Before the 1950's, \texttt{acts} in British English was primarily used as a legal term (with ordinances, enactments, laws etc). American English on the other hand used \texttt{acts} to refer to actions (as in \emph{acts of vandalism, acts of sabotage}). However in the $1960$'s British English started adopting the American usage. 

We hypothesize that this effect is observed due to globalization (the invention of radio, tv and the Internet), but leave a rigorous investigation of this phenomenon to future work. 

While our measure of semantic distance between languages does not capture lexical variation, introduction of new words etc, our work opens the door for future research to design better metrics for measuring semantic distances while also accounting for other forms of variation.

\section{Related Work}
Most of the related work can be organized into two areas: (a) Socio-variational linguistics (b) Word embeddings
\vspace{-0.1in}
\label{sec:related}
\paragraph{Socio-variational linguistics}
A large body of work studies how language varies according to geography and time \cite{eisenstein2010latent, eisenstein2011discovering,bamman2014gender,bamman2014distributed,kim-EtAl:2014:W14-25,kulkarni2015statistically, kenterad, gonccalves2014crowdsourcing}.

While previous work like \cite{gulordava-baroni:2011:GEMS,27806920,kim-EtAl:2014:W14-25, kenterad, brigadir2015analyzing} focus on temporal analysis of language variation, our work centers on methods to detect and analyze linguistic variation according to geography. A majority of these works also either restrict themselves to two time periods or do not outline methods to detect when changes are significant. Recently \cite{kulkarni2015statistically} proposed methods to detect statistically significant linguistic change over time that hinge on timeseries analysis. Since their methods explicitly model word evolution as a time series, their methods cannot be trivially applied to detect geographical variation. 

Several works on geographic variation \cite{bamman2014gender,eisenstein2010latent,o2010discovering,doyle2014mapping} focus on lexical variation. 
\citet{bamman2014gender} study lexical variation in social media like Twitter based on gender identity. 
\citet{eisenstein2010latent} describe a latent variable model to capture geographic lexical variation.
\citet{eisenstein2014diffusion} outline a model to capture diffusion of lexical variation in social media. Different from these studies, our work seeks to identify semantic changes in word meaning (usage) not limited to lexical variation. 
The work that is most closely related to ours is that of \citet{bamman2014distributed}. They propose a method to obtain geographically situated word embeddings and evaluate them on a semantic similarity task that seeks to identify words accounting for geographical location. Their evaluation typically focuses on named entities that are specific to geographic regions. Our work differs in several aspects: Unlike their work which does not explicitly seek to identify which words vary in semantics across regions, we propose methods to detect and identify which words vary across regions. While our work builds on their work to learn region specific word embeddings, we differentiate our work by proposing an appropriate null model, quantifying the change and assessing its significance. Furthermore our work is unique in the fact that we evaluate our method comprehensively on multiple web-scale datasets at different scales (both at a country level and state level).

 Measures of semantic distance have been developed for units of language (words, concepts etc) which \cite{mohammad2012distributional} provide an excellent survey. \citet{cooper2008measuring} study the problem of measuring semantic distance between languages, by attempting to capture the relative difficulty of translating various pairs of languages using bi-lingual dictionaries. Different from their work, we measure semantic distance between language dialects in an unsupervised manner (using word embeddings) and also analyze convergence patterns of language dialects over time.
\vspace{-0.1in}
\paragraph{Word Embeddings}
The concept of using distributed representations to learn a mapping from symbolic data to continuous space dates back to \citet{hinton1986learning}.
In a landmark paper, \citet{bengiolm} proposed a neural language model to learn word embeddings and demonstrated that they outperform traditional n-gram based models. \citet{mikolov2013efficient} proposed Skipgram models for learning word embeddings and demonstrated that they capture fine grained structures and linguistic regularities \cite{mikolov2013linguistic, phrases}.
Also \cite{perozzi2014inducing} induce language networks over word embeddings to reveal rich but varied community structure.
Finally these embeddings have been demonstrated to be useful features  for several NLP tasks \cite{senna,polyglot, polyglot-ner, chen2013expressive}.
\comment{
\subsection{Internet Linguistics}
Internet Linguistics seeks to study language use on media like online forums and social media, micro-blogs etc where Internet can significantly influence language use as well as electronic media like text messaging. There has been work on studying the language teens use while messaging and on online forums \cite{teenagers, ruin}. The phenomena of code-mixing which is prevalent to the multi-lingual nature of the web is an area of active research \cite{sharma2014borrowing,paolillo2011conversational,barman2014code}. \citet{Crystal:2011:ILS:2011923} provides an comprehensive survey of Internet Linguistics.}

\section{Conclusions}
\label{sec:Conclusions}
In this work, we proposed a new method to detect linguistic change across geographic regions.
Our method explicitly accounts for random variation, quantifying not only the change but also its significance.  
This allows for more precise detection than previous methods.

We comprehensively evaluate our method on large datasets at different levels of granularity -- from states in a country to countries spread across continents. 
Our methods are capable of detecting a rich set of changes attributed to word semantics, syntax, and code-mixing.
Using our method, we are able to characterize the semantic distances between dialectical variants over time. Specifically, we are able to observe the semantic convergence between British and American English over time, potentially an effect of globalization. This promising (although preliminary) result points to exciting research directions for future work.



\newpage
\section*{Acknowledgments}
\small
We thank David Bamman for sharing the code for training situated word embeddings. We thank Yingtao Tian for valuable comments.

\bibliographystyle{aaai}
\small
\bibliography{paper_icwsm}
\end{document}